\documentclass{article}

\usepackage[final]{corl_2026} 

\usepackage{booktabs}
\usepackage{graphicx} 
\usepackage[table]{xcolor}
\usepackage{pifont}
\usepackage{subcaption}
\usepackage{amsmath}
\usepackage{amssymb}
\usepackage{xspace}
\usepackage{enumitem}
\usepackage{titletoc}
\usepackage{wrapfig}
\usepackage[skins,breakable]{tcolorbox}

\definecolor{THUPurple}{rgb}{0.4314, 0.1686, 0.549}
\definecolor{amber}{rgb}{0.7490, 0.6235, 0.3529}
\definecolor{rose}{rgb}{0.6275, 0.4118, 0.5294}
\definecolor{sage}{rgb}{0.4941, 0.5843, 0.4078}
\definecolor{deemph}{gray}{0.6}
\definecolor{nacell}{HTML}{ECEBF0}
\definecolor{tasklavender}{HTML}{FBF8FE}
\definecolor{tasktitlelavender}{HTML}{A985BD}

\hypersetup{
  colorlinks = true,
  urlcolor = THUPurple,
  citecolor  = THUPurple,
  linkcolor = THUPurple,
  filecolor= THUPurple,
}

\newcommand{\cmarkx}{\textcolor{THUPurple}{\ding{55}}}
\newcommand{\method}{{\texttt{OpenHLM}}\xspace}
\makeatletter
\newif\ifshowtasklinenumbers
\if@conferencefinal
  \showtasklinenumbersfalse
\else
  \if@preprinttype
    \showtasklinenumbersfalse
  \else
    \showtasklinenumberstrue
  \fi
\fi
\newcommand{\tasktitlelinenumber}{%
  \ifshowtasklinenumbers
    \makebox[0pt][r]{%
      \normalfont\linenumberfont\textcolor{black}{\thelinenumber}\hspace{3.0em}%
    }%
    \global\advance\c@linenumber\@ne
  \fi
}
\makeatother
\newtcolorbox{taskbox}[1]{%
  enhanced,
  breakable,
  colback=tasklavender,
  colframe=THUPurple,
  colbacktitle=tasktitlelavender,
  coltitle=white,
  fonttitle=\bfseries,
  boxrule=0.6pt,
  arc=2mm,
  outer arc=2mm,
  left=1.5mm,
  right=1.5mm,
  top=1.2mm,
  bottom=1.2mm,
  before skip=0.8em,
  after skip=0.8em,
  before={\par\ifshowtasklinenumbers\nolinenumbers\fi},
  after={\par\ifshowtasklinenumbers\nolinenumbers\fi},
  before upper={\ifshowtasklinenumbers\internallinenumbers\addtolength{\linenumbersep}{8pt}\fi},
  after upper={\par},
  title={\tasktitlelinenumber\textbf{#1}}
}

\title{\method: An Empirical Recipe for Whole-Body Humanoid Loco-Manipulation}

\author{
  Yingdong Hu$^{1,2,3}${\hypersetup{linkcolor=black}\thanks{Core contributors}} \quad 
  Haodong Zhu$^{1}$\footnotemark[1] \quad 
  Boyuan Zheng$^{1,2}$\footnotemark[1] \quad 
  Yihang Hu$^{1,2}$\footnotemark[1] \quad 
  Tong Zhang$^{1,2,3}$\footnotemark[1]
  \\
  \textbf{Zunhao Chen}\textsuperscript{1}\;\; 
  \textbf{Junming Zhao}\textsuperscript{\text{1}}\;\; 
  \textbf{Ruiqian Nai}\textsuperscript{\text{1}}\;\; 
  \textbf{Yang Gao}$^{1,2,3}${\hypersetup{linkcolor=black}\thanks{Corresponding author}} \\
  \\
  \textsuperscript{1 }IIIS, Tsinghua University \;\;
  \textsuperscript{2 }Shanghai Qi Zhi Institute\;\;
  \textsuperscript{3 }Spirit.AI\;\;\\[0.3cm]
  \href{https://openhlm-project.github.io/}{https://openhlm-project.github.io/}
  \vspace{-0.7cm}
}

\begin{document}
\maketitle

\begin{abstract}
    Whole-body humanoid loco-manipulation requires coordinating the robot's entire kinematic chain. However, most existing systems typically decouple the upper and lower bodies into separate controllers, limiting such coordination and yielding behaviors similar to those of a wheeled dual-arm platform. In this paper, we ask what it takes to build a whole-body native vision-language-action (VLA) model that maps language and pixels directly to all of the humanoid's degrees of freedom. We conduct a systematic empirical study organized as a roadmap of one-variable-at-a-time experiments across three phases: whole-body teleoperation, VLA model design, and heterogeneous co-training. Our study yields several intriguing findings: a joint-based whole-body teleoperation interface outperforms alternatives that only partially expose the humanoid's degrees of freedom; a VLA pretrained on static and wheeled dual-arm platforms transfers surprisingly well to a humanoid's full action space; and co-training with HuMI, the humanoid analog of UMI, extends the policy to new objects and instructions without additional whole-body teleoperation on those targets. Following this roadmap yields \method, an open-source recipe for whole-body humanoid loco-manipulation. In a challenging long-horizon task that spans a wide vertical range of the humanoid, \method outperforms two state-of-the-art humanoid VLA baselines (GR00T N1.6 and $\Psi_0$) using less than half the total demonstration time. 
Our code, training data, and model checkpoints are available at \href{https://openhlm-project.github.io/}{https://openhlm-project.github.io/}.

\end{abstract}


\begin{figure}[h]
    \vspace{-1em}
    \centering
    \makebox[\linewidth][c]{%
    \includegraphics[width=.89\linewidth]{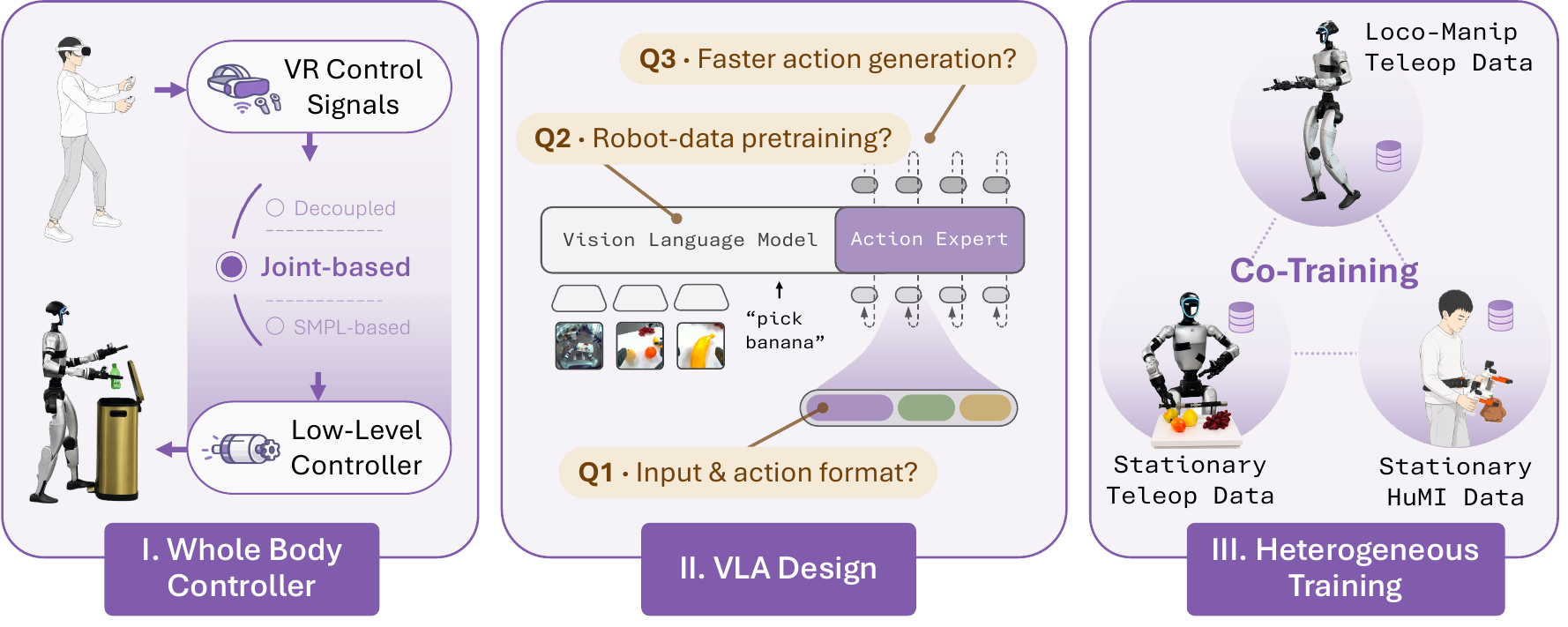}
    }
    \caption{\footnotesize \textbf{Overview of \method.} A roadmap of controlled experiments in three phases: \textbf{(I)} we compare teleop interfaces for a low-level whole-body controller and adopt a joint-based interface; \textbf{(II)} we adapt a manipulation VLA to the humanoid's full action space along several design axes; \textbf{(III)} we extend the policy to new objects and instructions by co-training the full loco-manipulation data with stationary teleop or HuMI~\cite{nai2026humanoid} demonstrations.}
    \label{fig:teaser}
    \vspace{-0.5em}
\end{figure}

\section{Introduction}
\label{sec:intro}

Humans perform complex loco-manipulation by coordinating their entire body, e.g., pressing a pedal with the foot, or squatting to reach a low shelf. 
Humanoid robots share a similar kinematic chain and, in principle, the same potential. However, most existing humanoid systems decouple the upper and lower bodies into separate controllers~\cite{cheng2024expressive,liu2024visual,lu2025mobile,ben2025homie,li2025amo,nvidia2025gr00tn16,wei2026psi0}.
Inverse kinematics drives the arms, a separate reinforcement-learning-trained controller drives the legs, and the two are stitched together through a navigation command and root-height signal.
This formulation limits whole-body coordination in two ways.
Visually, the motion is mechanical and unnatural.
Functionally, the lower body serves only as a mobile base, rather than an active participant in manipulation, leaving the humanoid closely akin to a wheeled dual-arm platform.
In response, a coordinated whole-body stack that reasons over the robot's entire kinematic chain has emerged as the path forward~\cite{ze2025twist2,figure2026helix02}, yet its design space remains largely unexplored.

A natural starting point is the two-level hierarchy that has recently emerged for such stacks~\cite{figure2026helix02}: a high-level vision-language-action (VLA) model~\cite{zitkovich2023rt} that maps language and pixels to whole-body commands, and a low-level controller that tracks them~\cite{he2024omnih2o,luo2025sonic}.
This decomposition raises three questions. The controller and the teleoperation interface built on it determine what demonstrations can be collected; how should they be designed? 
The VLA must handle a humanoid's full degrees of freedom, yet many widely used VLAs target static and wheeled dual-arm platforms~\cite{kim2024openvla,black2024pi0}; which adaptations actually matter?
And once the pipeline is established, whole-body teleop is too expensive to scale to every new object and instruction; can cheaper data sources fill the gap? 
This paper answers these questions through an empirical study, organized as a roadmap of controlled, one-variable-at-a-time experiments in three phases (Fig.~\ref{fig:teaser}): (1) whole-body controller and teleoperation (\S\ref{sec:roadmap-teleop}), (2) VLA model design (\S\ref{sec:roadmap-vla}), and (3) heterogeneous co-training (\S\ref{sec:roadmap-cotrain}). Within each phase, ablations on subsets of our HLM-12 benchmark quantify how each design choice affects task progress, building up to a concrete recipe for whole-body humanoid loco-manipulation.

Following this roadmap yields \textcolor{THUPurple}{\method} (\textcolor{THUPurple}{Open} \textcolor{THUPurple}{H}umanoid \textcolor{THUPurple}{L}oco-\textcolor{THUPurple}{M}anipulation), the open-source realization of this recipe; project materials are available at \href{https://openhlm-project.github.io/}{our project website}.
We highlight three of our surprising findings along the way as follows:
First, the teleoperation interface matters: a joint-level whole-body interface outperforms common alternatives such as VR 3-point control that partially expose the humanoid's degrees of freedom. 
Second, despite the large embodiment gap, a VLA pretrained on static and wheeled dual-arm platforms transfers surprisingly well to a humanoid's whole-body action space.
Third, co-training with cheaper manipulation-only sources, such as stationary teleop (feet planted, no locomotion) and HuMI~\cite{nai2026humanoid} (the humanoid analog of UMI~\cite{chi2024universal}), extends the policy to new objects and instructions without additional whole-body teleop on those targets.

Quantitatively, \method reaches 89\% average task progress on the 8 training tasks of our HLM-12 benchmark; on the 4 held-out tasks with no loco-manipulation teleop demos, co-training with cheaper data lifts task progress from 33\% to 87\%, closing most of the gap to the oracle (94\%).
On a long-horizon task spanning the humanoid's wide vertical workspace, \method outperforms two state-of-the-art humanoid VLAs (GR00T N1.6~\cite{nvidia2025gr00tn16} and $\Psi_0$~\cite{wei2026psi0}) at less than half the demonstration time; both baselines exhibit weak grasping and fail to track language-specified targets, despite including humanoid data in their pretraining.
To summarize, our main contributions are:

\begin{itemize}[leftmargin=4mm]
    \item \textbf{A systematic empirical study.} We explore the design space of whole-body humanoid loco-manipulation through controlled ablations across three phases (controller/teleoperation, VLA design, co-training), yielding a set of key findings about each design choice.
    \vspace{-0mm} \item \textbf{A concrete recipe for whole-body humanoid VLAs.} The study yields \method, an open-source recipe for VLAs that jointly control the humanoid's full action space. 
    The project website provides access to the codebase, checkpoints, and demonstration data.
    \vspace{-0mm} \item \textbf{Heterogeneous co-training for data efficiency.} We demonstrate that our system can be extended to new objects and instructions through cheaper data sources (stationary manipulation, HuMI~\cite{nai2026humanoid}) without additional loco-manipulation teleop, and characterize what each data stream supplies.
\end{itemize}

\section{Design Goals and Task Suite}
Before launching into the roadmap of \S\ref{sec:roadmap}, we first lay out the design goals the system aims to meet, and introduce the HLM-12 benchmark.

\textbf{Design goals.} We posit three desiderata for a physical humanoid loco-manipulation system.  Every subsection of the roadmap (\S\ref{sec:roadmap}) is framed as a response to one of them. 

\underline{\phantomsection\label{goal:G1}\textcolor{THUPurple}{(G1)} \textit{Whole-body native.}} One policy commands every joint of the humanoid simultaneously, treating the arms, knees, and feet as potential actuators for manipulation. Under the common decoupled formulation, the manipulation workspace shrinks to that of a wheeled dual-arm platform; behaviors that directly recruit the lower body fall outside the controller's expressible space.

\underline{\phantomsection\label{goal:G2}\textcolor{THUPurple}{(G2)} \textit{Language-steerable, and data-efficient per task.}} A single policy should drive the humanoid across many tasks, steered by a language prompt rather than by swapping checkpoints. Each new skill should be learnable from a modest number of demonstrations.

\underline{\phantomsection\label{goal:G3}\textcolor{THUPurple}{(G3)} \textit{Extensible through cheap data.}} Whole-body humanoid teleoperation is time-consuming and labor-intensive. The system should leverage cheaper heterogeneous data sources to reduce the demand for whole-body teleop, enabling faster skill extension to new objects and instructions.

\begin{figure}[b]
    \vspace{-1em}
    \centering
    \makebox[\linewidth][c]{%
    \includegraphics[width=1\linewidth]{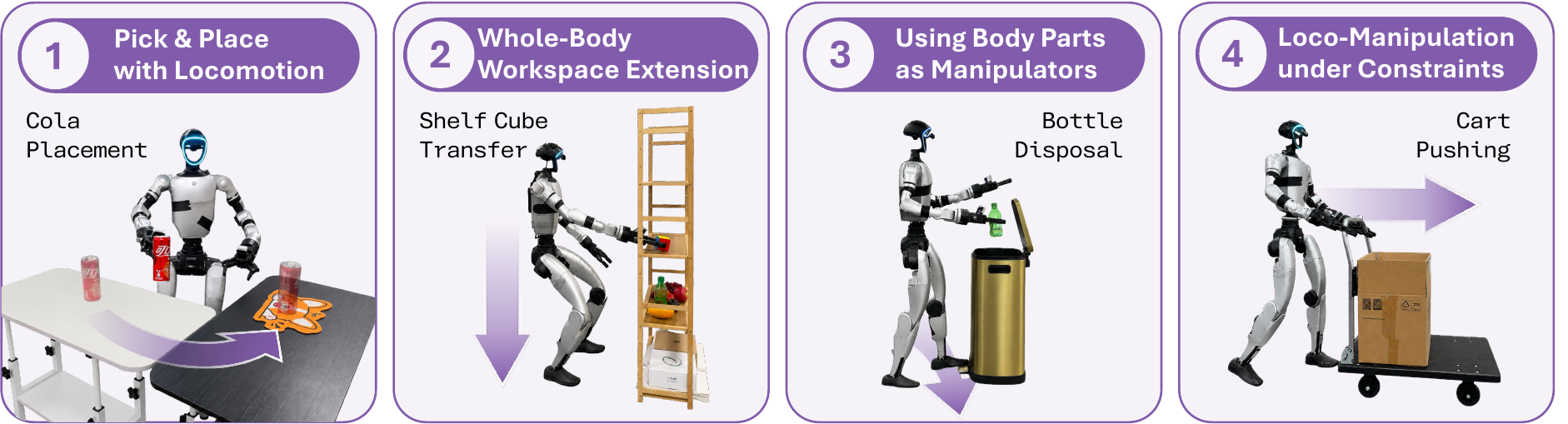}
    }
    \caption{\footnotesize \textbf{The HLM-12 Benchmark.} Tasks fall into four capability families targeting different aspects of whole-body behavior, with one representative per family shown. Full task specifications are in Appendix~\ref{appendix:12tasks}.}
    \label{fig:task_suite}
    \vspace{-1em}
\end{figure}

\textbf{The HLM-12 Benchmark.}
The HLM-12 benchmark contains 12 language-conditioned tasks, organized into four capability families that target different aspects of whole-body loco-manipulation behavior.
Fig.~\ref{fig:task_suite} illustrates a representative task from each family.
\underline{(1) \textit{Pick-and-place with locomotion.}} Tasks composing basic walking and grasping skills (e.g., \texttt{Cola Placement}) serve as basic checks of any loco-manipulation method.
\underline{(2) \textit{Whole-body workspace extension.}} Some target objects lie outside what upper-body articulation alone can cover; the humanoid must coordinate hip flexion, knee bend, and torso pitch with the arm to bring the end-effector into position (e.g., \texttt{Shelf Cube Transfer}).
\underline{(3) \textit{Using body parts as manipulators.}} Some tasks require body parts besides arms to serve as an end-effector (e.g., in \texttt{Bottle Disposal} the foot presses the pedal of the trash bin open). Such behaviors sit outside what a decoupled controller can express.
\underline{(4) \textit{Loco-manipulation under environmental constraint.}} 
Object geometry (e.g., \texttt{Sword Extraction} requires pulling along the sword axis) or contact requirements (e.g., \texttt{Cart Pushing} requires a firm grasping posture while walking) may restrict viable loco-manipulation motions.
A policy's performance on each of the 12 tasks is evaluated by its average progress over 5 independent rollouts, as task progress captures more nuanced failure modes than a binary success rate.
We present the full specification of all 12 tasks and more about the evaluation protocol in Appendix~\ref{appendix:12tasks}.

\section{Building a Whole-Body Loco-Manipulation System: A Roadmap}
\label{sec:roadmap}

We construct the system through controlled experiments
in three phases that answer the questions raised in \S\ref{sec:intro}.
\textit{Controller and teleoperation} (\S\ref{sec:roadmap-teleop}): how to design the controller and its teleop interface for high-quality whole-body demonstrations.
\textit{VLA design} (\S\ref{sec:roadmap-vla}): which adaptations turn a VLA built for static and wheeled robots into a whole-body humanoid policy.
\textit{Heterogeneous co-training} (\S\ref{sec:roadmap-cotrain}): whether cheaper data sources can extend the policy beyond what whole-body teleop alone covers.

\subsection{Low-Level Controller \& Teleoperation}
\label{sec:roadmap-teleop}

We follow a \textit{two-level hierarchical control framework}, in line with recent humanoid loco-manipulation stacks~\cite{nvidia2025gr00tn16,shi2026egohumanoid,wei2026psi0,figure2026helix02}. 
A high-level policy 
(human teleoperator or trained VLA)
takes vision and language as input and emits reference whole-body commands at low frequency (typically 10\,Hz). 
A lightweight low-level controller consumes these commands and outputs target joint positions at higher frequency (typically 50\,Hz), which are then tracked by a PD controller. With this framework fixed, two design questions immediately follow.

\textbf{Opening question.} \textit{What should the interface between the high-level policy and the low-level controller look like, and what properties should the low-level controller itself satisfy?}

We study the teleop interface along two axes.
First, \textit{expressivity}: interfaces exposing only a subset of the humanoid's degrees of freedom make certain tasks unreachable by construction.
Second, \textit{demonstration quality}: interfaces with similar expressivity can still differ in the quality of demonstrations they elicit, directly affecting the learned policy.
As for the low-level controller, we study how preview latency (i.e., how far the controller sees future motion) affects final performance.

\textbf{Joint-based whole-body teleoperation beats decoupled control and VR 3-point.} We compare three teleoperation interfaces commonly used in recent humanoid loco-manipulation systems, with implementation details provided in Appendix~\ref{appendix:teleop}:

\underline{(1) \textit{Decoupled control teleoperation.}} 
The upper and lower bodies are two decoupled systems. Operator-provided targets (head and the two wrists) are mapped to upper-body joints through inverse kinematics, while an RL-trained lower-body controller, conditioned on these upper-body commands, tracks a base-velocity and root-height command. This formulation is widely adopted by recent systems including AMO~\cite{li2025amo}, $\Psi_0$~\cite{wei2026psi0}, and GR00T N1.5/N1.6~\cite{nvidia2025gr00tn16}; we use the GR00T variant~\cite{nvidia2026decoupledwbc} here.

\underline{(2) \textit{VR 3-point teleoperation.}}
A widely used scheme in humanoid teleoperation~\cite{he2024omnih2o,luo2025sonic}; we adopt the SONIC variant~\cite{luo2025sonic} here. The operator supplies head and wrist poses via a VR headset, with a navigation command from its joystick. A learned kinematic motion planner produces the lower-body motion, yielding a hybrid command of three upper-body keypoints and lower-body joint positions that is tracked by the SONIC controller.

\underline{(3) \textit{Joint-based whole-body teleoperation.}}
A portable motion-capture rig (here, a PICO4U VR suit~\cite{pico2024pico4ultra}) captures the operator's whole-body motion and retargets it in real time to \textit{every humanoid joint} via GMR~\cite{araujo2025retargeting}. 
The resulting joint trajectory is tracked by a general motion tracker (SONIC).

We select three tasks stressing different capabilities, collect 40 demonstrations for each task and teleop method, and train one VLA per teleop method (VLA details in \S\ref{sec:roadmap-vla}).
Results in Table~\ref{tab:teleop} suggest that
\textit{Joint-based whole-body teleoperation} is the only interface that completes all three tasks, reaching 80\%--87\% task progress. 
\textit{Decoupled control} walks in small, visibly unnatural steps: on \texttt{Cola Placement}, it averages 42.3 footsteps, about 3.5 times the joint-based policy's 12.0, and it cannot perform \texttt{Bottle Disposal} because the controller cannot command the required foot motion.
\textit{VR 3-point teleoperation} can only perform \texttt{Cola Placement}, where the policy often stalls in front of the can and reaches 40\% progress over a 67.8\,s rollout; tasks requiring foot depressing or kneeling are out of reach.
Besides, it is visually observable that the joint-based method produces much more fluent motion and gentler gaits than the other two methods, leading to a much lower frequency of hardware failure. Table~\ref{tab:vibration} provides quantitative evidence consistent with this phenomenon. 
Based on all these results, \textit{we adopt joint-based whole-body teleoperation as the data-collection interface.}

\begin{table}[h]
\centering
\resizebox{\columnwidth}{!}{
\begin{tabular}{lccc|ccc|ccc}
\toprule
& \multicolumn{3}{c}{\texttt{Cola Placement}} 
& \multicolumn{3}{c}{\texttt{Shelf Cup Transfer}} 
& \multicolumn{3}{c}{\texttt{Bottle Disposal}} \\
\cmidrule(lr){2-4} \cmidrule(lr){5-7} \cmidrule(lr){8-10}
Teleop method 
& Prog. (\%) & Time (s) & Steps 
& Prog. (\%) & Time (s) & Steps 
& Prog. (\%) & Time (s) & Steps \\
\midrule
Decoupled control 
& $66.7{\scriptstyle\,\pm\,14.9}$ & 36.7 & 42.3 
& $93.3{\scriptstyle\,\pm\,6.7}$ & 33.6 & 38.4 
& \multicolumn{3}{>{\columncolor{nacell}}c}{\cmarkx} \\

VR 3-point teleoperation
& $40.0{\scriptstyle\,\pm\,6.7}$ & 67.8 & 12.3
& \multicolumn{3}{>{\columncolor{nacell}}c|}{\cmarkx}
& \multicolumn{3}{>{\columncolor{nacell}}c}{\cmarkx} \\

Joint-based whole-body teleop.
& $86.7{\scriptstyle\,\pm\,8.2}$ & 40.2 & 12.0 
& $80.0{\scriptstyle\,\pm\,13.3}$ & 41.8 & 10.5 
& $85.0{\scriptstyle\,\pm\,6.1}$ & 29.7 & 10.3 \\
\bottomrule
\end{tabular}
}
\vspace{0.2em}
\caption{\footnotesize
  \textbf{Teleop method comparison.}
  \textbf{Prog. (\%)}, mean task progress over 5 rollouts;
  \textbf{Time~(s)}, mean rollout duration;
  \textbf{Steps}, mean footsteps per rollout.
  A purple \cmarkx~marks a task the method cannot perform by construction:
  \texttt{Shelf Cup Transfer} requires squatting;
  \texttt{Bottle Disposal} requires the foot to depress a pedal.
}
\label{tab:teleop}
\vspace{-0.6em}
\end{table}

\begin{wraptable}{r}{0.6\textwidth}
    \centering
    \small
    \vspace{-0.8em}
    \setlength{\tabcolsep}{3pt}
    \resizebox{\linewidth}{!}{%
      \label{tab:tilt-rate}
  \begin{tabular}{lccc}
    \hline
    Task & Joint-based & VR three-point & Decoupled \\
    \hline
    \texttt{Cola Placement} & \textbf{0.187} & 0.235 & 0.439 \\
    \texttt{Shelf Cup Transfer} & \textbf{0.178} & -- & 0.424 \\
    \hline
  \end{tabular}
    }
    \caption{Average IMU vibration amplitudes ($A=\text{sqrt}(\text{avg.}(\dot{\phi}^2+\dot{\theta}^2))$) in the collected demonstration dataset, evaluated above three methods across two tasks.}
    
    \vspace{-0.4em}
    \label{tab:vibration}
\end{wraptable}

\begin{wrapfigure}{r}{0.6\textwidth}
    \centering
    \vspace{-1.4em}
    \includegraphics[width=\linewidth]{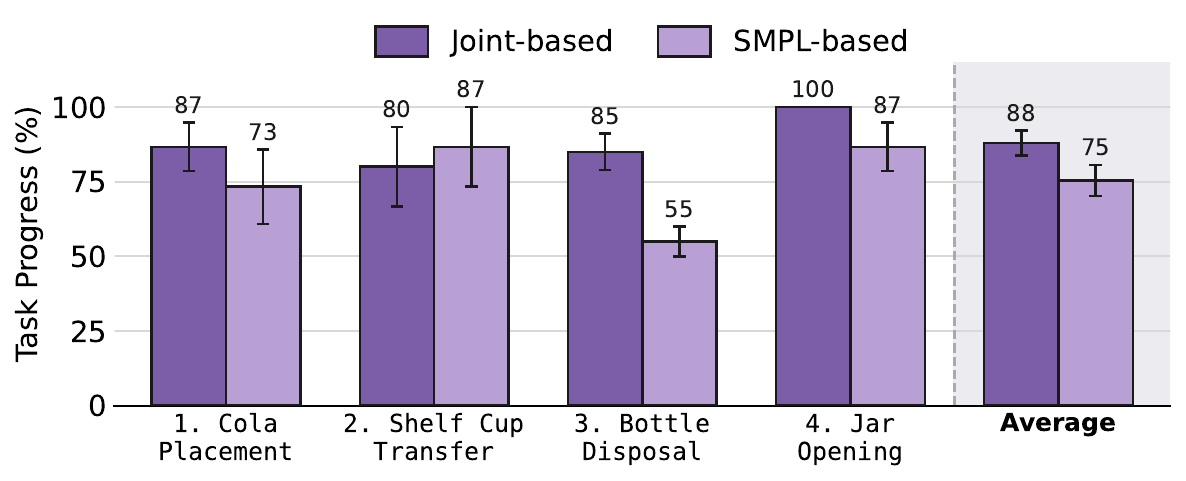}
    \vspace{-1.5em}
    \caption{\footnotesize \textbf{Joint-based vs.\ SMPL whole-body teleop.}}
    \label{fig:smpl}
    \vspace{-0.7em}
\end{wrapfigure}

\textbf{Joint-space retargeting beats native SMPL recording.}
SMPL~\cite{loper2015smpl} is a natural representation of human whole-body motion.
Using SMPL as the action representation
skips the online retargeting step and thus eliminates errors from an imperfect retargeter, 
but gives the VLA a much higher-dimensional target space (81-D vs.\ 32-D for joint-space).
We collect the same number of demos under both motion representations on our 4-task subset; comparison results are shown in Fig.~\ref{fig:smpl}.
The joint-based policy reaches 88\% average progress while the SMPL-based policy reaches 75\%.
SMPL introduces failure modes that do not appear in the joint-space policy.
For example, on \texttt{Bottle Disposal}, the SMPL policy often lifts the heel without sufficiently lifting the toes, leaving inadequate clearance to depress the pedal.
On \texttt{Cola Placement}, it sometimes walks too close to the table and knocks over the can; neither failure appears in the collected demonstrations.
We attribute the gap to the redundancy of the 81-D SMPL action space compared with the 32-D robot-joint space (Appendix~\ref{appendix:teleop}).
Based on this result, \textit{we retarget whole-body demonstrations to robot joint space online during data collection.}

\textbf{Future-frame preview latency: 0.2\,s reaches a good balance.}
The preview latency $\Delta t$ decides how far the low-level controller can see the future reference motion. Longer preview may yield smoother motion but increases teleoperation delay.
We sweep $\Delta t \in \{0, 0.2, 0.4, 0.6\}$\,s on task \texttt{Cola Placement} and find that our default choice $\Delta t = 0.2$\,s reaches a good balance between locomotion fluency and manipulation accuracy. 
(See detailed result \& analysis in Appendix~\ref{appendix:latency_gap}.)
Our data-collection pipeline for all later phases is now settled: \textit{joint-based whole-body teleoperation, retargeted to robot joint space online, with 0.2\,s preview latency.}

\subsection{Whole-Body VLA Policy Design}
\label{sec:roadmap-vla}

We now turn to the high-level policy. An appealing starting point is a pretrained VLA that already brings vision-language reasoning and manipulation priors. 
However, existing VLAs nearly all target static or wheeled dual-arm platforms; none were designed for humanoid loco-manipulation.

\textbf{Opening question.} \textit{How do we adapt a VLA pretrained on static and wheeled dual-arm platforms into a whole-body humanoid policy, and which design choices actually matter?}

\begin{figure}[h]
    \vspace{-1em}
    \centering
    \includegraphics[width=1.0\linewidth]{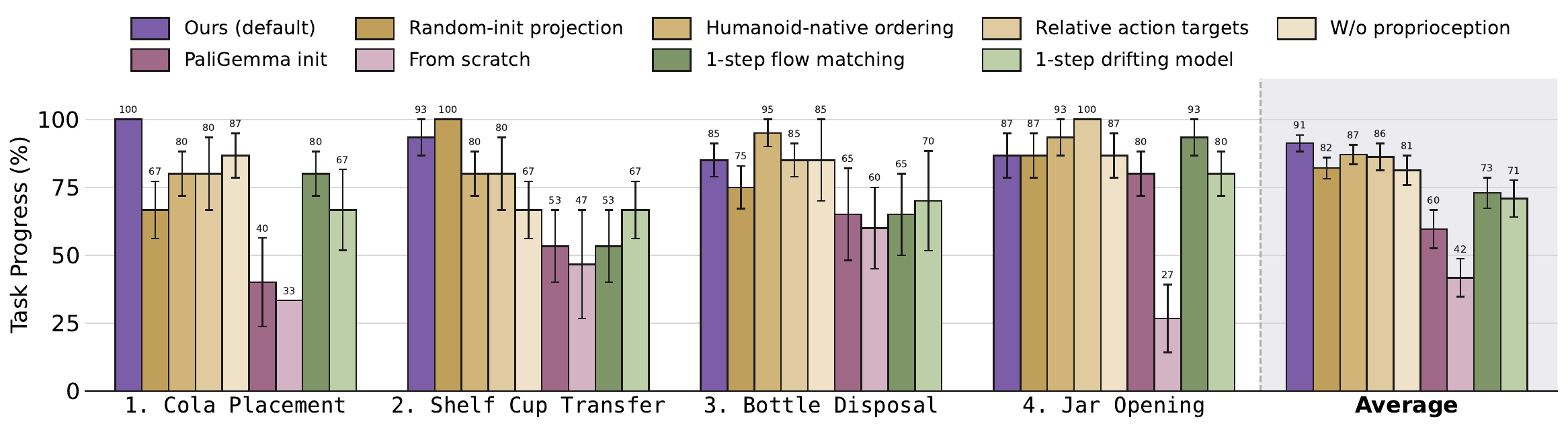}
    \vspace{-1.8em}
    \caption{\footnotesize \textbf{VLA design ablations on the 4-task subset.} \textcolor{amber}{Amber}: interface ablations (one choice flipped per bar); performance gaps are minor. \textcolor{rose}{Rose}: pretraining ablations; robot pretraining ($\pi_{0.5}$) dominates. \textcolor{sage}{Sage}: one-step action generation; both underperform the 10-step baseline.}
    \label{fig:vla_results}
    \vspace{-1em}
\end{figure}

We organize the exploration into three families: (1) \textit{Action and proprioception interface} — adapting the VLA's original lower-DoF action space to the humanoid's high-DoF commands.
(2) \textit{The role of pretraining} — whether robot pretraining ($\pi_{0.5}$~\cite{intelligence2025pi_}) is needed, or vision-language pretraining alone (PaliGemma~\cite{beyer2024paligemma}) or even training from scratch can already work. 
(3) \textit{Faster action generation} — whether the multi-step flow matching~\cite{lipman2022flow} can be replaced by single-step inference. We fix a default configuration and ablate one component at a time on the 4-task subset; results are shown in Fig.~\ref{fig:vla_results}. 

\textbf{Action and proprioception interface: adaptations barely affect performance.}
We use $\pi_{0.5}$~\cite{intelligence2025pi_} as our default backbone and study the interface between the VLA and the humanoid. We ablate four design choices around its input/output format.
\underline{(1) \textit{Action projection initialization.}} $\pi_{0.5}$'s action projection supports up to 32 action dimensions, needing to be resized to 34 (32 dims from \S\ref{sec:roadmap-teleop} plus two grippers) to fit our humanoid. We compare random re-initialization against weight surgery (default), which preserves the pretrained weights for the first 32 dimensions in the input and output linear projection layers and randomly initializes only the new entries.
\underline{(2) \textit{Action ordering.}} $\pi_{0.5}$'s pretrained action vector is laid out as $[\text{left arm}, \text{left gripper}, \text{right arm}, \text{right gripper}]$. We can preserve this layout and append the humanoid-specific waist and leg joints at the end (default), or pick a fresh humanoid-native ordering (e.g., legs first).
\underline{(3) \textit{Absolute vs.\ relative action targets.}} Predict absolute joint positions (default), or relative deltas with each action chunk re-expressed relative to the first action in that chunk.
\underline{(4) \textit{w./w.o. Proprioceptive input.}} Besides visual and language inputs, whether to feed the full current joint-position vector as input to the policy (default) or not.

We ablate each choice independently, holding the other three at default; results appear as the \textcolor{amber}{amber group} in Fig.~\ref{fig:vla_results}. All variations 
produce only slight drops in 4-task average task progress, with no qualitative shift in rollout behavior or failure modes. 
The small numerical drops are plausibly attributed to lower robustness or just evaluation noise. 
None of these four choices is, in itself, the bottleneck of humanoid VLA adaptation. 
However, the choices cannot necessarily be combined arbitrarily: in an additional experiment, removing proprioception and switching to relative actions together caused the policy to drift into out-of-distribution states and fail catastrophically.
Since no single alternative beats the default, we keep the default for the rest of the paper:
\textit{we adopt weight surgery on the action projection, the pretrained bimanual ordering, absolute joint targets, and proprioception as input.}

\textbf{Pretraining on non-humanoid robots contributes to humanoid VLA.}
With the interface fixed, we ask whether pretraining on non-humanoid robot data (static and wheeled dual-arm) transfers to humanoid despite the substantial embodiment gap, or whether a vision-language backbone alone suffices. We compare three backbone initializations: (i) $\pi_{0.5}$ (non-humanoid robot-pretrained); (ii) PaliGemma (same architecture, no robot data); (iii) random initialization. Results appear as the \textcolor{rose}{rose group} in Fig.~\ref{fig:vla_results}: the $\pi_{0.5}$-initialized version reaches 91\% average task progress while the other two drop to 60\% and 42\%.
Compared to the $\pi_{0.5}$ version, PaliGemma is consistently weaker at grasping and rarely recovers from a failed grasp, and random-initialization only learns a rough gait with poor manipulation ability. These differences show that $\pi_{0.5}$'s manipulation prior, especially the closed-loop ``see error, correct, retry'' behavior transfers despite the embodiment gap. 
Despite this on-robot gap, validation action mean squared error (MSE) is nearly indistinguishable between the $\pi_{0.5}$- and PaliGemma-initialized models throughout fine-tuning, showing that action MSE alone is a poor proxy for real-world performance.
The cross-embodiment gap from dual-arm pretraining to our humanoid is real, but dwarfed by the gap between any robot pretraining and none at all. \textit{We initialize from $\pi_{0.5}$ for all subsequent experiments.}

\textbf{Faster action generation: one-step alternatives underperform.}
At inference, $\pi_{0.5}$'s flow-matching~\cite{lipman2022flow} action head integrates the learned vector field over 10 denoising steps. A single-step alternative would cut inference latency from $\sim$90\,ms to $\sim$60\,ms on our 5080 GPU, potentially improving closed-loop responsiveness. We test two alternatives: \textit{one-step flow matching} and \textit{drifting model}~\cite{deng2026generative} (Appendix~\ref{appendix:vla}).
Results (\textcolor{sage}{sage group} in Fig.~\ref{fig:vla_results}) show that both alternatives underperform the baseline by roughly 20 task-progress points, even though their validation action MSE is lower ($\sim$0.007 versus $\sim$0.009 for the 10-step baseline).
We hypothesize that single-step inference produces actions 
jitterier and less temporally smooth on the robot.
\textit{We keep multi-step flow matching for action generation.}

\textbf{Whole-body teleop scaling: 40 demonstrations is a data-efficient default.}
Throughout the paper, we use 40 whole-body teleop demonstrations per task as the default  budget. We conduct experiments to prove that this amount is appropriate for the difficulty of our tasks. See details in Appendix~\ref{appendix:scaling}.

At this point we have a humanoid-adapted VLA: a $\pi_{0.5}$-initialized backbone with weight-surgery action projection, the pretrained bimanual ordering, absolute joint targets, proprioception as input, and multi-step flow matching inference. Carrying it to the 8 training tasks (40 demonstrations each), the system reaches \textit{89\%} average task progress (Fig.~\ref{fig:co-training}), confirming that the choices made on the 4-task subset generalize to the broader training distribution.

\subsection{Heterogeneous Co-Training}
\label{sec:roadmap-cotrain}

Scaling our whole-body VLA to every new task through loco-manipulation teleop is expensive. We turn to ask whether and how cheaper data sources can introduce new abilities through co-training. 
We focus on two: (i) stationary same-embodiment teleoperation (feet-planted manipulation, no locomotion); (ii) HuMI-collected robot-free demonstrations~\cite{nai2026humanoid} (captured with low-cost wearable devices). We measure co-training on the 4 held-out tasks that whole-body teleop never covers.

\textbf{Opening question.} \textit{Can cheaper data extend the VLA's capability to new tasks, and if so, how?}

\begin{figure}[h]
    \centering
    \includegraphics[width=0.84\linewidth]{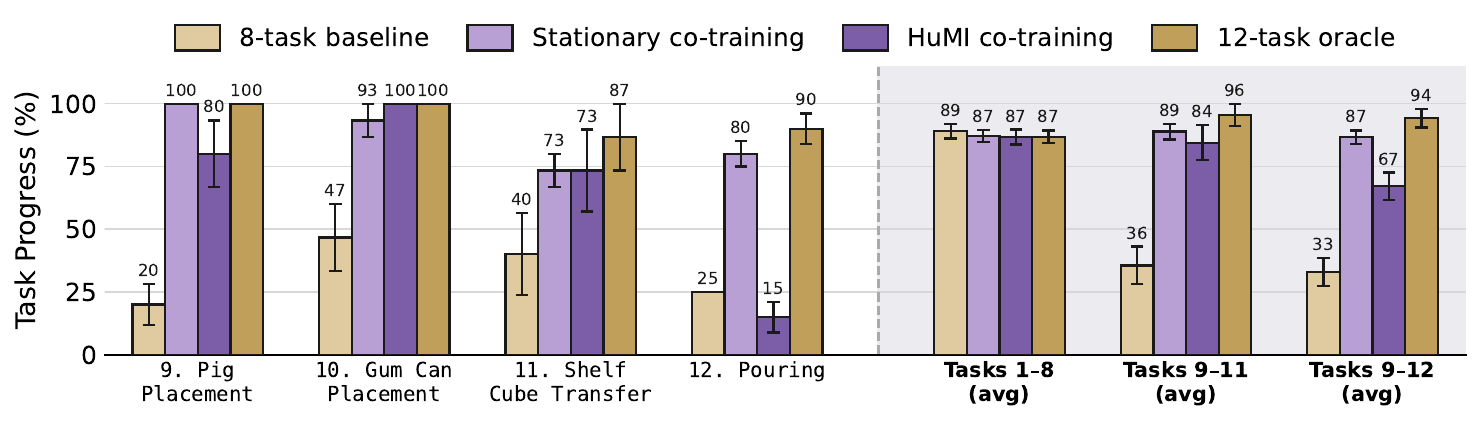}
    \vspace{-0.6em}
    \caption{\footnotesize \textbf{Heterogeneous co-training results.} Progress on the 4 held-out tasks and averages over training tasks (1--8), motion-reuse held-out tasks (9--11), and all held-out tasks (9--12). Full breakdown in Appendix~\ref{appendix:per_task}.}
    \vspace{-0.3em}
    \label{fig:co-training}
\end{figure}

\textbf{Cheaper stationary co-training extends the policy to held-out tasks.} 
Stationary teleoperation operates the same humanoid with feet planted in place: manipulation only, no locomotion. 
The motivation is straightforward: manipulation usually demands higher positional precision than locomotion, and thus requires relatively more demonstrations. Besides, manipulation-only demos are much cheaper to collect (shorter episodes, smaller scenes, faster resets). 
 The average duration of 40 demos for each held-out task is 13\,min under stationary teleop v.s. 21\,min under full teleop, excluding overheating shutdowns and scene resets, which take much longer under full teleop.

We co-train the VLA on whole-body teleop for the 8 training tasks and stationary teleop for the 4 held-out tasks, and compare against two references: the 8-task baseline (whole-body teleop on 8 tasks only) and the 12-task oracle (whole-body teleop on all 12 tasks). Fig.~\ref{fig:co-training} shows that stationary co-training 
maintains performance on 
the 8 training tasks; on held-out tasks, it lifts average progress from 33\% (8-task baseline) to 87\%, closing much of the gap to the oracle. 
Tasks 9--11 (\texttt{Pig Placement}, \texttt{Gum Can Placement}, and \texttt{Shelf Cube Transfer}) reuse motions from the training tasks but introduce new objects and language prompts, whereas \texttt{Pouring} (Task 12) requires a vessel-tilt motion absent from the 8-task whole-body teleop set.
Stationary co-training approaches the oracle on both groups, indicating that the cheaper demonstrations can provide both new semantic grounding and new motions.
These results suggest that \textit{cheaper stationary teleop, co-trained with the whole-body set, extends the policy to held-out tasks by supplying both semantic grounding and new motions.}

\textbf{HuMI co-training delivers new semantic understanding but not new motions.} HuMI~\cite{nai2026humanoid} uses two hand-held UMI~\cite{chi2024universal} grippers and three body-pose trackers~\cite{vive_ultimate_tracker} to capture SE(3) keypoint trajectories, which IK lifts to full-body joint positions. The resulting action vector matches the joint-based whole-body teleop interface of \S\ref{sec:roadmap-teleop}, allowing co-training without architectural changes. As with stationary teleop, we collect manipulation only. 
Averaged over held-out tasks, 40 HuMI demonstrations total only 7\,min, compared with 13\,min for stationary teleoperation.

We co-train the 8-task whole-body teleop set with HuMI on the 4 held-out tasks (40 demos each). Fig.~\ref{fig:co-training} shows no degradation on the 8 training tasks. 
On tasks 9--11 (motion-reuse, new objects and prompts), HuMI matches stationary co-training, grounding new semantics. 
On \texttt{Pouring} (task 12), HuMI fails to acquire the new vessel-tilt motion absent from the 8-task training set.
We attribute this to two domain gaps between HuMI and teleop. Visually, the two differ in cameras (RealSense vs.\ rectified GoPro fisheye) and grippers (humanoid's adaptive parallel vs.\ HuMI's rigid hand-held).
As for motion, HuMI trajectories are retargeted to robot motions through inverse kinematics, and these motions lie outside the distribution of humanoid teleop motions.
At the current data scale, these gaps make HuMI act primarily as semantic rather than motion supervision; scaling HuMI data may enable motion transfer, which we leave to future work.
\textit{With the current data budget, HuMI co-training delivers new semantic understanding but not new motions, at roughly half the operator-time cost of stationary same-embodiment teleop.}

The roadmap is complete and yields \method. Joint-based whole-body teleoperation (\S\ref{sec:roadmap-teleop}) and a $\pi_{0.5}$-initialized humanoid-adapted VLA (\S\ref{sec:roadmap-vla}) together cover the humanoid's full degrees of freedom and reach 89\% average task progress on the 8 training tasks, meeting \hyperref[goal:G1]{\textcolor{THUPurple}{G1}} and \hyperref[goal:G2]{\textcolor{THUPurple}{G2}}. Heterogeneous co-training with cheaper data sources (\S\ref{sec:roadmap-cotrain}) extends the policy to new objects, instructions, and motions without additional whole-body teleop, meeting \hyperref[goal:G3]{\textcolor{THUPurple}{G3}}. Comparisons so far have been internal; \S\ref{sec:system} benchmarks \method against state-of-the-art humanoid VLAs on a long-horizon task.

\section{Long-Horizon System Comparison}
\label{sec:system}

\begin{wrapfigure}{r}{0.6\textwidth}
    \centering
    \vspace{-1.5em}
    \includegraphics[width=\linewidth,trim=178bp 60bp 40bp 200bp,clip]{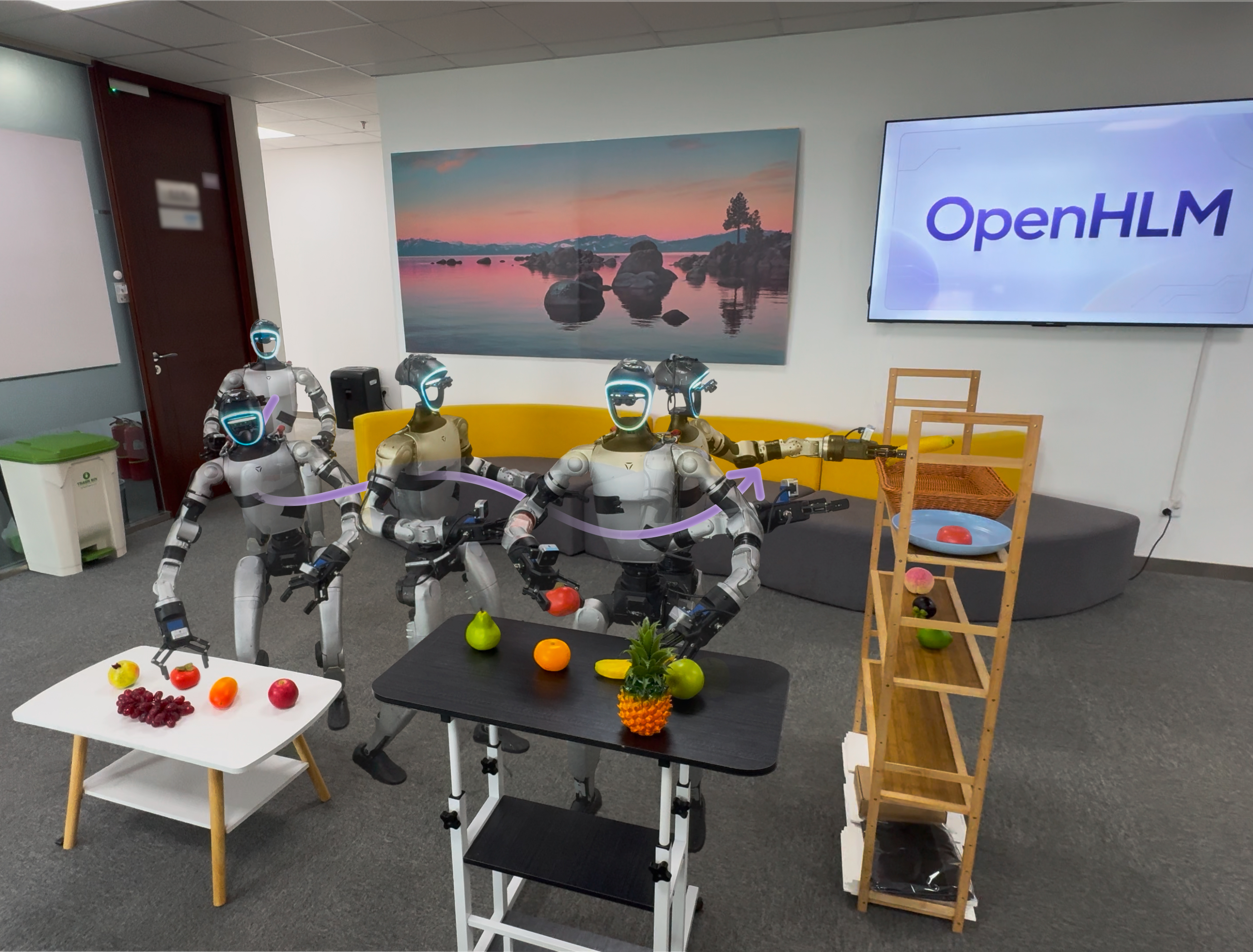}
    \vspace{-1.3em}
    \caption{\footnotesize \textbf{Long-horizon task.}}
    \label{fig:long_horizon}
    \vspace{-1.5em}
\end{wrapfigure}

\textbf{Baseline systems.} We compare \method\,{\scriptsize (HuMI co-training)} against two representative humanoid VLAs, GR00T N1.6~\cite{nvidia2025gr00tn16} and $\Psi_0$~\cite{wei2026psi0}. Both use decoupled low-level control; implementation details are in Appendix~\ref{appendix:baselines}.

\textbf{Task \& data.}
The humanoid performs a long-horizon language-conditioned task spanning its large vertical workspace (Fig.~\ref{fig:long_horizon}). From a home pose, it walks to a low coffee table and picks up \{fruit 1\} with the right hand, walks to a medium-height table and picks up \{fruit 2\} with the left hand, then walks to a tall shelf and places them in separate containers on the top shelf. The instruction specifies \{fruit 1\} and \{fruit 2\} from five fruits (banana, peach, mangosteen, lemon, tomato), with distractors at each site, giving $P(5,2)=20$ ordered pairs.
We collect 6 demonstrations per pair under each condition. 
GR00T N1.6 and $\Psi_0$ are trained with full loco-manipulation teleop on all 20 pairs (120 demos). 
\method\,{\scriptsize (HuMI co-training)} is trained with full loco-manipulation teleop only on the $P(3,2)=6$ pairs drawn from \{banana, peach, mangosteen\} (36 demos), and with HuMI demonstrations of the manipulation phases for the remaining 14 pairs that include lemon or tomato (84 demos); this tests whether HuMI co-training extends the policy to new objects without further whole-body teleop.
\method\,{\scriptsize (teleop oracle)} instead uses whole-body teleop on all 20 pairs and serves as the oracle.

\begin{wraptable}{r}{0.5\textwidth}
    \centering
    \small
    \vspace{-0.4em}
    \setlength{\tabcolsep}{3pt}
    \resizebox{\linewidth}{!}{%
    \begin{tabular}{@{}lcc@{}}
        \toprule
        Method{\,\scriptsize (long-horizon task)} & Progress (\%) & Demo Dur. \\
        \midrule
        $\Psi_0$ & $48.8{\scriptstyle\,\pm\,4.4}$ & $2.70$\,h \\
        GR00T N1.6 & $57.5{\scriptstyle\,\pm\,4.6}$ & $2.70$\,h \\
        \method\,{\scriptsize (HuMI co-training)} & $87.5{\scriptstyle\,\pm\,3.7}$ & $1.14$\,h \\
        \textcolor{deemph}{\method\,{\scriptsize (teleop oracle)}} & \textcolor{deemph}{$97.5{\scriptstyle\,\pm\,1.7}$} & \textcolor{deemph}{$2.73$\,h} \\
        \bottomrule
    \end{tabular}%
    }
    \caption{\textbf{Task progress \& demo duration.}
    }
    \vspace{-0.7em}
    \label{tab:demo_time}
\end{wraptable}

\textbf{Results.}
Table~\ref{tab:demo_time} reports task progress and total demo duration (see detailed evaluation protocol in Appendix~\ref{appendix:long_horizon}). \method{\scriptsize (HuMI co-training)} reaches $87.5\%$ task progress, far above $\Psi_0$ ($48.8\%$) and GR00T N1.6 ($57.5\%$) and within $10$ points of the teleop oracle ($97.5\%$), at less than half the operator time ($\sim$1.1 vs.\ $\sim$2.7 hours). 
Both baselines exhibit weak grasping: they execute a stereotyped arm trajectory rather than tracking the language-specified fruit. $\Psi_0$ additionally stops short of the tall shelf, indicating weaker locomotion. 
Notably, GR00T N1.6 and $\Psi_0$ both include Unitree G1 demonstrations in their pretraining data, while \method's $\pi_{0.5}$ initialization does not. 
This suggests that mixing humanoid data into pretraining is not enough on its own; strong humanoid VLAs still depend on getting the design details in \S\ref{sec:roadmap} right.

\section{Related Work}

\textbf{Whole-body humanoid teleoperation.}
Teleoperation is a foundational data source for humanoid loco-manipulation, and the interface itself shapes what behaviors the trained policy can express. \textit{Decoupled control} drives the lower body with a separate locomotion controller that takes velocity or coarse motion commands~\cite{cheng2024expressive,liu2024visual,lu2025mobile,ben2025homie,li2025amo,nvidia2025gr00tn16,wei2026psi0}; this rules out coordinated whole-body behaviors and any use of the lower body as a manipulator. \textit{Whole-body teleoperation} instead commands all degrees of freedom jointly~\cite{ze2025twist,fu2024humanplus,he2024omnih2o,zhang2025hub}, typically through a learned motion-tracking controller~\cite{liao2025beyondmimic,zeng2025behavior,chen2025gmt}. Within this paradigm, TWIST2~\cite{ze2025twist2} establishes the data-collection side, but pairs it with single-task visuomotor imitation~\cite{chi2025diffusion} rather than a multi-task VLA; SONIC~\cite{luo2025sonic} scales motion-tracking controllers to unprecedented size, but reports only a handful of experiments in which the controller is plugged into a VLA. How the teleoperation interface interacts with a VLA-driven policy is therefore left open. To this end, we use whole-body teleop as the data source for a VLA and systematically compare alternative teleop formats, offering practical insights for future use (\S\ref{sec:roadmap-teleop}).

\textbf{Humanoid VLAs.} 
Vision-language-action models fine-tune a vision-language backbone to emit robot actions. RT-2~\cite{zitkovich2023rt}, OpenVLA~\cite{kim2024openvla}, and the $\pi$ series~\cite{black2024pi0,intelligence2025pi_,intelligence2025pi,intelligence2026pi} establish strong generalization across manipulation scenes, but on fixed-base or wheeled platforms whose action distribution is dominated by arm and wrist motion. Two recent efforts extend this recipe to humanoids. GR00T N1.6~\cite{nvidia2025gr00tn16} pretrains a vision-language model with a DiT~\cite{peebles2023scalable} action head on a mixture that includes humanoid demonstrations; $\Psi_0$~\cite{wei2026psi0} pretrains on large-scale egocentric human video~\cite{hoque2025egodex} and then mid-trains an action expert on humanoid demonstrations. 
Both follow the same recipe: add humanoid data to pretraining, then finetune. Yet in our system-level comparison (\S\ref{sec:system}), both underperform a $\pi_{0.5}$-initialized humanoid-adapted VLA (\S\ref{sec:roadmap-vla}) that pretrains on no humanoid data at all. The bottleneck instead lies in design details: how to humanoid-adapt a VLA backbone, what teleop format to fit, and how to mix in cheap data. These are the questions \S\ref{sec:roadmap} works through.

\textbf{Heterogeneous data for humanoids.} 
Cheap, robot-free data collection has a rich toolkit: handheld sensorized grippers~\cite{chi2024universal,liu2024fastumi,lin2025data}, wearable smart glasses~\cite{engel2023project}, and AR/VR headsets~\cite{apple_vision_pro,meta_quest_3,pico2024pico4ultra}. Together, these tools have powered scene-understanding pretraining~\cite{grauman2024ego,grauman2022ego4d}, visual representation learning~\cite{nair2022r3m,ma2022vip}, motion priors~\cite{yang2025egovla}, latent action codebooks~\cite{ye2025latent,bu2025univla}, and direct co-training of robot policies~\cite{yuan2025motiontrans,qiu2025humanoid}. 
However, most of this work targets fixed-base or wheeled manipulation; humanoid loco-manipulation is still emerging. As a first step in this direction, HuMI~\cite{nai2026humanoid} extends UMI with wearable body trackers and IK retargeting to full-body humanoid joints; we adopt it as a starting point, while the broader space (smart glasses, AR/VR headsets) remains open. The closest concurrent effort, EgoHumanoid~\cite{shi2026egohumanoid}, also studies human-to-humanoid co-training, but along an orthogonal axis: it pairs human and teleop demonstrations on the same full tasks to drive environment generalization, whereas \S\ref{sec:roadmap-cotrain} uses cheap data streams to extend the policy to new objects and language prompts that whole-body teleop never covers.

\section{Discussion, Conclusion and Limitations}

In this work, we study the design choices behind reliable whole-body humanoid loco-manipulation: how to teleoperate the humanoid, adapt a manipulation VLA to the humanoid's full action space, and use cheap data beyond full teleoperation.
Our roadmap (\S\ref{sec:roadmap}) studies these questions with controlled experiments and yields \method: joint-based whole-body teleoperation, a $\pi_{0.5}$-initialized humanoid-adapted VLA, and heterogeneous co-training for low-cost object and language expansion.
On a challenging long-horizon task that interleaves multiple locomotion and manipulation stages, \method outperforms GR00T N1.6 and $\Psi_0$ with less than half of their demo duration.
Our code, data, and models are available through the project website to support future humanoid robotics research.

Our work has several limitations that future research can address. 
(1) Though our work explores a broad design space of humanoid VLAs, it is still confined to our limited experiment resources (e.g., we only use Unitree G1 as the humanoid platform). Whether our conclusions hold in more general settings (e.g., other humanoid robots, more diverse scenes) remains uncertain. (2) Heterogeneous co-training does not yet close the gap to whole-body teleoperation demonstrations.  HuMI data improves grounding for new objects and language prompts but is less effective at teaching novel motion patterns, likely due to visual and action gaps between robot-free and on-robot demonstrations, or our limited data budget.  How to better utilize cheap robot-free data remains an open question. (3) We do not explore VLA architecture design specifically for loco-manipulation tasks, and during training on the current architecture, several findings remain mechanistically underexplained: validation action MSE does not reliably predict on-robot performance, and one-step action generation underperforms despite lower MSE.
We hope that future researchers could extend our roadmap following these directions.

\section*{Core Contributors}
The five core contributors jointly shaped this work, each leading complementary and often overlapping parts of the project. The descriptions below highlight each author's primary areas of leadership, and should not be interpreted as disjoint partitions or as a ranking of individual effort.

\begin{description}
    \item[\textbf{Yingdong Hu.}] Led the manuscript outline and high-level policy (VLA) design, and contributed to the low-level controller design and the teleoperation and evaluation software.
    \item[\textbf{Haodong Zhu.}] Led the teleoperation and evaluation software design and the organization of the full codebase, and contributed to teleoperation data collection and baseline implementation.
    \item[\textbf{Boyuan Zheng.}] Led co-training data collection and baseline implementation, including data collection and evaluation, and contributed to teleoperation data collection.
    \item[\textbf{Yihang Hu.}] Led hardware design and assembly, task scene construction and process design, and teleoperation data collection, and contributed to the teleoperation and evaluation software.
    \item[\textbf{Tong Zhang.}] Led the low-level controller design and testing, and contributed to high-level policy (VLA) design and hardware design and assembly.
\end{description}

\clearpage
\acknowledgments{This research was conducted with the support of the Shanghai Qi Zhi Institute \& Spirit AI Innovation Program. Funding and support for this work were also provided by the Tsinghua University - Keystone Electrical (Zhejiang) Co.,Ltd Joint Research Center for Embodied Multimodal Artificial Intelligence (JCEMAI). Additionally, we would like to extend our thanks to the Xiongan AI Institute. This project is also support by Beijing Natural Science Foundation(JR25019).}

\bibliography{example}  

\clearpage
\newpage

\appendix
\section*{Appendices}
\startcontents[appendices]
\printcontents[appendices]{l}{0}{\setcounter{tocdepth}{2}}

\newpage
\section{The HLM-12 Benchmark}
\label{appendix:12tasks}

\subsection{The 12 Tasks}

We present the detailed settings of our HLM-12 benchmark. For each task, we provide the corresponding \textbf{Language Prompt}, \textbf{Detailed Process}, \textbf{Evaluation Rubric}, and the \textbf{Capability Stressed}.

\ifshowtasklinenumbers
\begin{nolinenumbers}
\fi

\begin{taskbox}{Task 1: \texttt{Cola Placement}}

        \begin{itemize}[leftmargin=4mm]
            \item \textbf{Language Prompt:} Walk forward to the table, put the red can on the orange mouse pad with the right hand, turn around and walk back.
        
            \item \textbf{Detailed Process:} The robot needs to walk to an appropriate position in front of the white table, locate the cola can placed randomly among other distractor objects, and carefully grasp it without pushing it away or causing it to fall. It then turns about 90 degrees to the left using its feet and waist to face the mouse pad, and places the can onto it. Finally, the robot turns around and walks back to its initial location.
        
            \item \textbf{Evaluation Rubric:} 3 points in total:
            \begin{itemize}
                \item The robot successfully walks to the correct position in front of the table and later returns to its initial location. (1 pt)
                \item The robot picks up the correct object (the cola can). (1 pt)
                \item The cola can is placed exactly on the target location (the orange mouse pad). (1 pt)
            \end{itemize}

            \item \textbf{Capability Stressed:} \textit{Pick-and-place with locomotion}.
        \end{itemize}

        \par\nolinenumbers

        \begin{center}
            \includegraphics[width=0.98\linewidth]{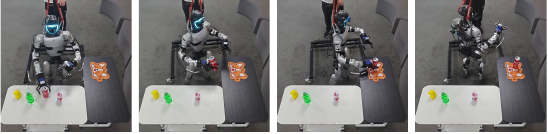}
        \end{center}
\end{taskbox}

\begin{taskbox}{Task 2: \texttt{Shelf Cup Transfer}}

        \begin{itemize}[leftmargin=4mm]
            \item \textbf{Language Prompt:} Walk forward to the shelf, put the white cup from the upper level to the lower level with the right hand, turn around and walk back.
        
            \item \textbf{Detailed Process:} The robot needs to walk to an appropriate position in front of the shelf, locate and grasp the white cup placed randomly among other distractor objects on the upper level, then bend down and carefully place it onto the lower level of the shelf. Finally, the robot turns around and walks back to its initial location.
        
            \item \textbf{Evaluation Rubric:} 3 points in total:
            \begin{itemize}
                \item The robot successfully walks to the correct position in front of the shelf and later returns to its initial location. (1 pt)
                \item The robot picks up the correct object (the white cup). (1 pt)
                \item The white cup is successfully placed on the lower level of the shelf. (1 pt)
            \end{itemize}

            \item \textbf{Capability Stressed:} \textit{Whole-body workspace extension}.
        \end{itemize}

        \par\nolinenumbers

        \begin{center}
            \includegraphics[width=0.98\linewidth]{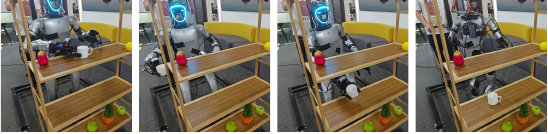}
        \end{center}
\end{taskbox}

\begin{taskbox}{Task 3: \texttt{Bottle Disposal}}

        \begin{itemize}[leftmargin=4mm]
            \item \textbf{Language Prompt:} Walk forward to the dust bin, throw the green bottle into the dust bin, turn around and walk back.
        
            \item \textbf{Detailed Process:} The robot needs to walk to an appropriate position in front of the dust bin, locate and grasp the green bottle among other distractor objects on the table to the right, then press the pedal with the left foot to open the dust bin lid, and place the bottle completely inside the bin. Finally, the robot turns around and walks back to its initial location.
        
            \item \textbf{Evaluation Rubric:} 4 points in total:
            \begin{itemize}
                \item The robot successfully walks to the correct position in front of the dust bin and later returns to its initial location. (1 pt)
                \item The robot picks up the correct object (the green bottle). (1 pt)
                \item The robot successfully presses the pedal to keep the lid open. (1 pt)
                \item The bottle is successfully placed inside the bin. (1 pt)
            \end{itemize}

            \item \textbf{Capability Stressed:} \textit{Using body parts as manipulators}.
        \end{itemize}

        \par\nolinenumbers

        \begin{center}
            \includegraphics[width=0.98\linewidth]{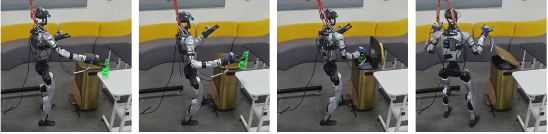}
        \end{center}
\end{taskbox}

\begin{taskbox}{Task 4: \texttt{Jar Opening}}

        \begin{itemize}[leftmargin=4mm]
            \item \textbf{Language Prompt:} Walk forward to the white table, open the red jar with both hands, turn around and walk back.
        
            \item \textbf{Detailed Process:} The robot needs to walk to an appropriate position in front of the white table, locate the red jar placed randomly among other distractor objects, grasp the upper lid with the right hand and the lower body with the left hand, then separate the two parts by pulling them apart with both hands. After opening the jar, the robot places both parts back onto the table. Finally, the robot turns around and walks back to its initial location.
        
            \item \textbf{Evaluation Rubric:} 3 points in total:
            \begin{itemize}
                \item The robot successfully walks to the correct position in front of the table and later returns to its initial location. (1 pt)
                \item The robot uses the right hand to grasp the correct part of the correct object (the upper lid of the red jar). (1 pt)
                \item The robot uses the left hand to grasp the correct part (the lower body of the red jar), and the jar is successfully opened. (1 pt)
            \end{itemize}

            \item \textbf{Capability Stressed:} \textit{Loco-manipulation under environmental constraint}.
        \end{itemize}

        \par\nolinenumbers

        \begin{center}
            \includegraphics[width=0.98\linewidth]{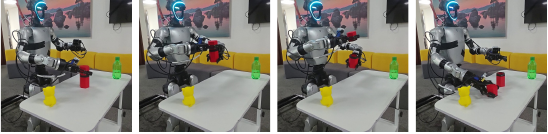}
        \end{center}
\end{taskbox}

\begin{taskbox}{Task 5: \texttt{Toy Stowing}}

        \begin{itemize}[leftmargin=4mm]
            \item \textbf{Language Prompt:} Walk forward to the drawer, pick up the blue toy bear with the left hand, open the drawer with the right hand and drop the blue toy bear inside, close the drawer with the right leg, turn around and walk back. 
        
           \item \textbf{Detailed Process:} The robot needs to walk to an appropriate position in front of the drawer, locate and grasp the blue toy bear among other distractor objects with the left gripper, then close the right gripper, insert it into the small gap of the drawer, and pull the drawer open. After placing the toy bear inside the drawer, the robot uses the right leg to gently push the drawer closed. Finally, the robot turns around and walks back to its initial location.
        
        \item \textbf{Evaluation Rubric:} 5 points in total:
        \begin{itemize}
            \item The robot successfully walks to the correct position in front of the drawer and later returns to its initial location. (1 pt)
            \item The robot picks up the correct object (the blue toy bear). (1 pt)
            \item The robot successfully inserts the right gripper into the gap and pulls the drawer open. (1 pt)
            \item The toy bear is placed exactly inside the drawer. (1 pt)
            \item The drawer is gently pushed closed using the right leg. (1 pt)
        \end{itemize}
        
        \item \textbf{Capability Stressed:} \textit{Using body parts as manipulators}.
        \end{itemize}

        \par\nolinenumbers

        \begin{center}
            \includegraphics[width=0.98\linewidth]{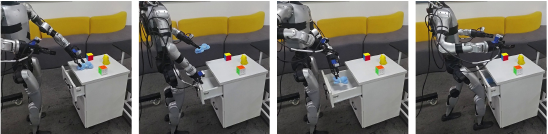}
        \end{center}
\end{taskbox}

\begin{taskbox}{Task 6: \texttt{Sword Extraction}}

        \begin{itemize}[leftmargin=4mm]
            \item \textbf{Language Prompt:} Walk forward to the white drawer, pull out the sword with both hands, turn around and walk back.
        
            \item \textbf{Detailed Process:} The robot needs to walk to an appropriate position in front of the white cabinet, locate the sword on the rack, grasp the hilt with the right hand and lift the sword, then grasp the scabbard with the left hand and pull the hilt and scabbard apart with both hands to extract the sword. Afterward, the robot turns around and walks back to its initial location while holding the sword.
        
            \item \textbf{Evaluation Rubric:} 4 points in total:
            \begin{itemize}
                \item The robot successfully walks to the correct position in front of the cabinet and later returns to its initial location. (1 pt)
                \item The robot uses the right hand to grasp the correct part of the correct object (the hilt of the sword). (1 pt)
                \item The robot uses the left hand to grasp the correct part (the scabbard of the sword). (1 pt)
                \item The robot successfully pulls the sword out of the scabbard. (1 pt)
            \end{itemize}
    
            \item \textbf{Capability Stressed:} \textit{Loco-manipulation under environmental constraint}.
        \end{itemize}

        \par\nolinenumbers

        \begin{center}
            \includegraphics[width=0.98\linewidth]{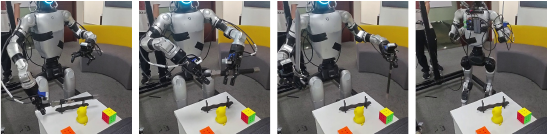}
        \end{center}
\end{taskbox}

\begin{taskbox}{Task 7: \texttt{Cart Pushing}}

        \begin{itemize}[leftmargin=4mm]
            \item \textbf{Language Prompt:} Walk forward to the cart, throw the box onto the cart, then push the cart away.
        
            \item \textbf{Detailed Process:} The robot needs to walk to an appropriate position in front of the cart, locate and grasp the box on the table to the right, and drop it into the large box on the cart. It then grasps the left and right sides of the cart handle sequentially with the corresponding hands. Afterward, the robot pushes the cart forward to carry the contents away.
        
            \item \textbf{Evaluation Rubric:} 4 points in total:
            \begin{itemize}
                \item The robot successfully walks to the correct position in front of the cart and then successfully pushes the cart forward. (1 pt)
                \item The robot uses the right hand to grasp the correct object (the yellow iron box). (1 pt)
                \item The iron box is placed exactly into the large box on the cart. (1 pt)
                \item The robot successfully grasps both sides of the cart handle. (1 pt)
            \end{itemize}
    
            \item \textbf{Capability Stressed:} \textit{Loco-manipulation under environmental constraint}.
        \end{itemize}

        \par\nolinenumbers

        \begin{center}
            \includegraphics[width=0.98\linewidth]{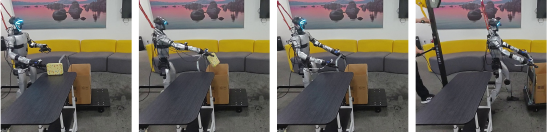}
        \end{center}
\end{taskbox}

\begin{taskbox}{Task 8: \texttt{Shuttlecock Setup}}

        \begin{itemize}[leftmargin=4mm]
            \item \textbf{Language Prompt:} Walk forward to the black table, pull out one shuttle and put it on the racket with the left hand, turn around and walk back.
        
            \item \textbf{Detailed Process:} The robot needs to walk to an appropriate position in front of the black table, locate the upside-down shuttlecock dispenser placed on the tabletop, grasp the bottom shuttlecock with the left gripper and pull it downward to remove it, then place it onto the racket face on the left side. Finally, the robot turns around and walks back to its initial location.
        
            \item \textbf{Evaluation Rubric:} 3 points in total:
            \begin{itemize}
                \item The robot successfully walks to the correct position in front of the table and later returns to its initial location. (1 pt)
                \item The robot uses the left hand to locate the shuttlecock dispenser and pull out one shuttlecock successfully. (1 pt)
                \item The robot places the shuttlecock exactly onto the racket face. (1 pt)
            \end{itemize}

            \item \textbf{Capability Stressed:} \textit{Loco-manipulation under environmental constraint}.
        \end{itemize}

        \par\nolinenumbers

        \begin{center}
            \includegraphics[width=0.98\linewidth]{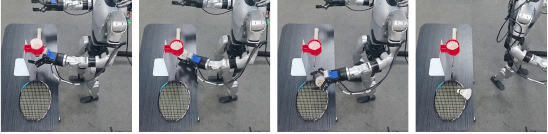}
        \end{center}
\end{taskbox}

\begin{taskbox}{Task 9: \texttt{Pig Placement}}

        \begin{itemize}[leftmargin=4mm]
            \item \textbf{Language Prompt:} Walk forward to the table, put the yellow toy pig on the blue mouse pad with the right hand, turn around and walk back.
        
            \item \textbf{Detailed Process:} The robot needs to walk to an appropriate position in front of the white table, locate the yellow toy pig placed randomly among other distractor objects, and carefully grasp it. It then turns about 90 degrees to the left using its feet and waist to face the mouse pad, and places the toy pig onto it. Finally, the robot turns around and walks back to its initial location.
        
            \item \textbf{Evaluation Rubric:} 3 points in total:
            \begin{itemize}
                \item The robot successfully walks to the correct position in front of the table and later returns to its initial location. (1 pt)
                \item The robot picks up the correct object (the yellow toy pig). (1 pt)
                \item The toy pig is placed exactly on the target location (the blue mouse pad). (1 pt)
            \end{itemize}
    
            \item \textbf{Capability Stressed:} \textit{Pick-and-place with locomotion}.
        \end{itemize}

        \par\nolinenumbers

        \begin{center}
            \includegraphics[width=0.98\linewidth]{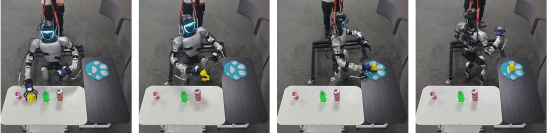}
        \end{center}
\end{taskbox}
    
\begin{taskbox}{Task 10: \texttt{Gum Can Placement}}
    
        \begin{itemize}[leftmargin=4mm]
            \item \textbf{Language Prompt:} Walk forward to the table, put the pink gum can on the maroon mouse pad with the right hand, turn around and walk back.
        
            \item \textbf{Detailed Process:} The robot needs to walk to an appropriate position in front of the white table, locate the pink gum can placed randomly among other distractor objects, and carefully grasp it. It then turns about 90 degrees to the left using its feet and waist to face the mouse pad, and places the gum can onto it. Finally, the robot turns around and walks back to its initial location.
        
            \item \textbf{Evaluation Rubric:} 3 points in total:
            \begin{itemize}
                \item The robot successfully walks to the correct position in front of the table and later returns to its initial location. (1 pt)
                \item The robot picks up the correct object (the pink gum can). (1 pt)
                \item The gum can is placed exactly on the target location (the maroon mouse pad). (1 pt)
            \end{itemize}
    
            \item \textbf{Capability Stressed:} \textit{Pick-and-place with locomotion}.
        \end{itemize}

        \par\nolinenumbers

        \begin{center}
            \includegraphics[width=0.98\linewidth]{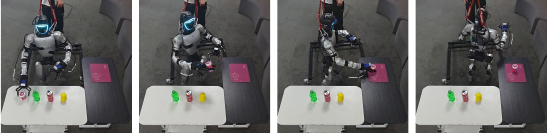}
        \end{center}
\end{taskbox}

\newpage

\begin{taskbox}{Task 11: \texttt{Shelf Cube Transfer}}

        \begin{itemize}[leftmargin=4mm]
            \item \textbf{Language Prompt:} Walk forward to the shelf, put the cube from the upper level to the lower level with the right hand, turn around and walk back.
        
            \item \textbf{Detailed Process:} The robot needs to walk to an appropriate position in front of the shelf, locate and grasp the cube placed randomly among other distractor objects on the upper level, then bend down and carefully place it onto the lower level of the shelf. Finally, the robot turns around and walks back to its initial location.
        
            \item \textbf{Evaluation Rubric:} 3 points in total:
            \begin{itemize}
                \item The robot successfully walks to the correct position in front of the shelf and later returns to its initial location. (1 pt)
                \item The robot picks up the correct object (the cube). (1 pt)
                \item The cube is successfully placed on the lower level of the shelf. (1 pt)
            \end{itemize}

            \item \textbf{Capability Stressed:} \textit{Whole-body workspace extension}.
        \end{itemize}

        \par\nolinenumbers

        \begin{center}
            \includegraphics[width=0.98\linewidth]{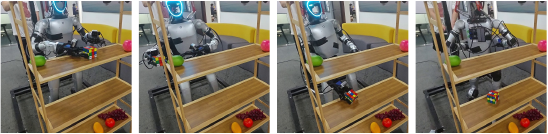}
        \end{center}
\end{taskbox}

\begin{taskbox}{Task 12: \texttt{Pouring}}

        \begin{itemize}[leftmargin=4mm]
            \item \textbf{Language Prompt:} Walk forward to the white table, pour the contents from the blue cup into the white plate with the right hand, turn around and walk back.
        
            \item \textbf{Detailed Process:} The robot needs to walk to an appropriate position in front of the white table, locate the blue cup placed randomly among other distractor objects, grasp it with the right hand, and pour the dice inside into the white plate placed on the towel at the center of the table. The robot then places the empty cup onto the left side of the table while turning around to the left, and finally walks back to its initial location.
        
            \item \textbf{Evaluation Rubric:} 4 points in total:
            \begin{itemize}
                \item The robot successfully walks to the correct position in front of the table and later returns to its initial location. (1 pt)
                \item The robot uses the right hand to grasp the correct object (the blue cup). (1 pt)
                \item The robot pours the dices successfully into the white plate. (1 pt)
                \item The robot places the empty cup on the left side of the table. (1 pt)
            \end{itemize}

            \item \textbf{Capability Stressed:} \textit{Loco-manipulation under environmental constraint}.
        \end{itemize}

        \par\nolinenumbers

        \begin{center}
            \includegraphics[width=0.98\linewidth]{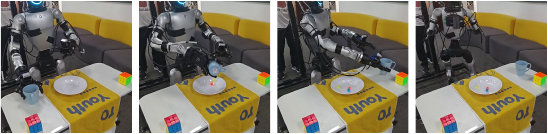}
        \end{center}
\end{taskbox}

\ifshowtasklinenumbers
\end{nolinenumbers}
\linenumbers
\fi

\subsection{Overall Evaluation Protocol}
We adopt a shared, rigorous evaluation protocol: 
Every (policy, task) pair is evaluated in the real world over five independent rollouts. Note that we use five rollouts rather than more because each loco-manipulation trial requires resetting both the scene and the robot to its home pose, making it substantially slower than a stationary-manipulation trial. 
Across the five rollouts the target object is placed in different positions, and each rollout introduces a different layout of distractor objects.
For each task, the same five initial scene configurations (recorded by photographs) are used across all policies to ensure fair comparisons.
Rather than recording only binary success/failures, we record task progress by assigning points based on the evaluation rubrics of each task introduced before. The final score of a specific rollout appears as a task progress fraction in $[0,1]$. 
For example, in the \texttt{Cola Placement} task, if the robot does everything well except that the cola can is placed outside the mouse pad, it loses 1 point and gets $2/3=67\%$ task progress.
Compared with the binary success rate, task progress captures more nuanced failure modes. 
We report standard errors alongside the mean.

\section{Hardware Setup}
\label{appendix:hardware}

\subsection{Humanoid Robot}
\label{appendix:humanoid}

\begin{figure}[h!]
    \centering
    \begin{minipage}[t]{0.48\linewidth}
        \centering
        \includegraphics[width=\linewidth]{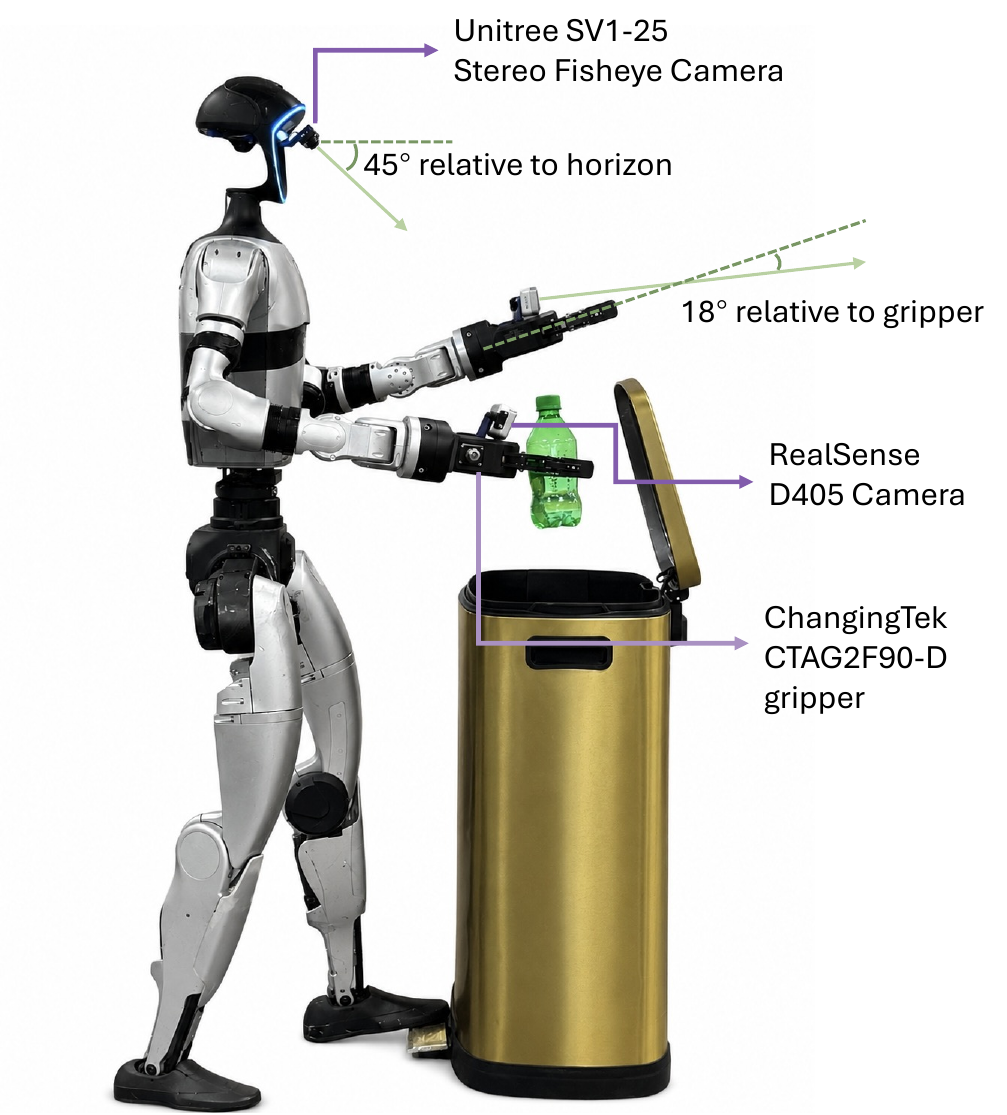}
        \caption{\textbf{Humanoid robot hardware.} The Unitree G1 is equipped with wrist-mounted grippers and onboard cameras.}
        \label{fig:hardware_setup}
    \end{minipage}
    \hfill
    \begin{minipage}[t]{0.46\linewidth}
        \centering
        \includegraphics[width=0.74\linewidth]{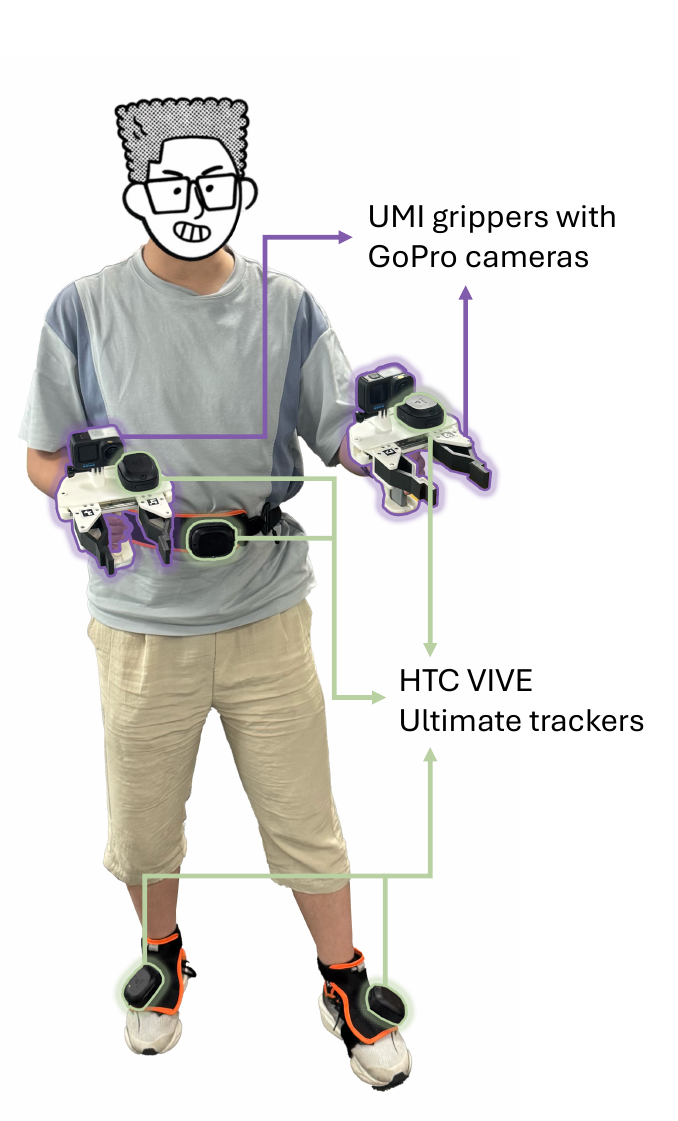}
        \caption{\textbf{HuMI hardware.} Handheld grippers and body trackers are used to collect teleoperation-free demonstrations.}
        \label{fig:humi_hardware}
    \end{minipage}
\end{figure}

All our loco-manipulation tasks are carried out by a Unitree G1 robot, with two ChangingTek CTAG2F90-D grippers~\cite{changingtek_ctag2f90d} equipped on its wrists to enable grasping.
Visual perception is provided by a Unitree SV1-25 fisheye stereo camera mounted on the robot's head, together with two Intel RealSense D405 cameras mounted on the wrists. The associated connection cables are routed and secured along the back of the humanoid to avoid interfering motions. Figure~\ref{fig:hardware_setup} provides detailed illustrations of the hardware configuration.

\subsection{Whole-Body Teleoperation Hardware}

\begin{figure}[h!]
    \vspace{-1em}
    \centering
    \begin{minipage}[c]{0.56\linewidth}
        \centering
        \includegraphics[height=0.32\textheight]{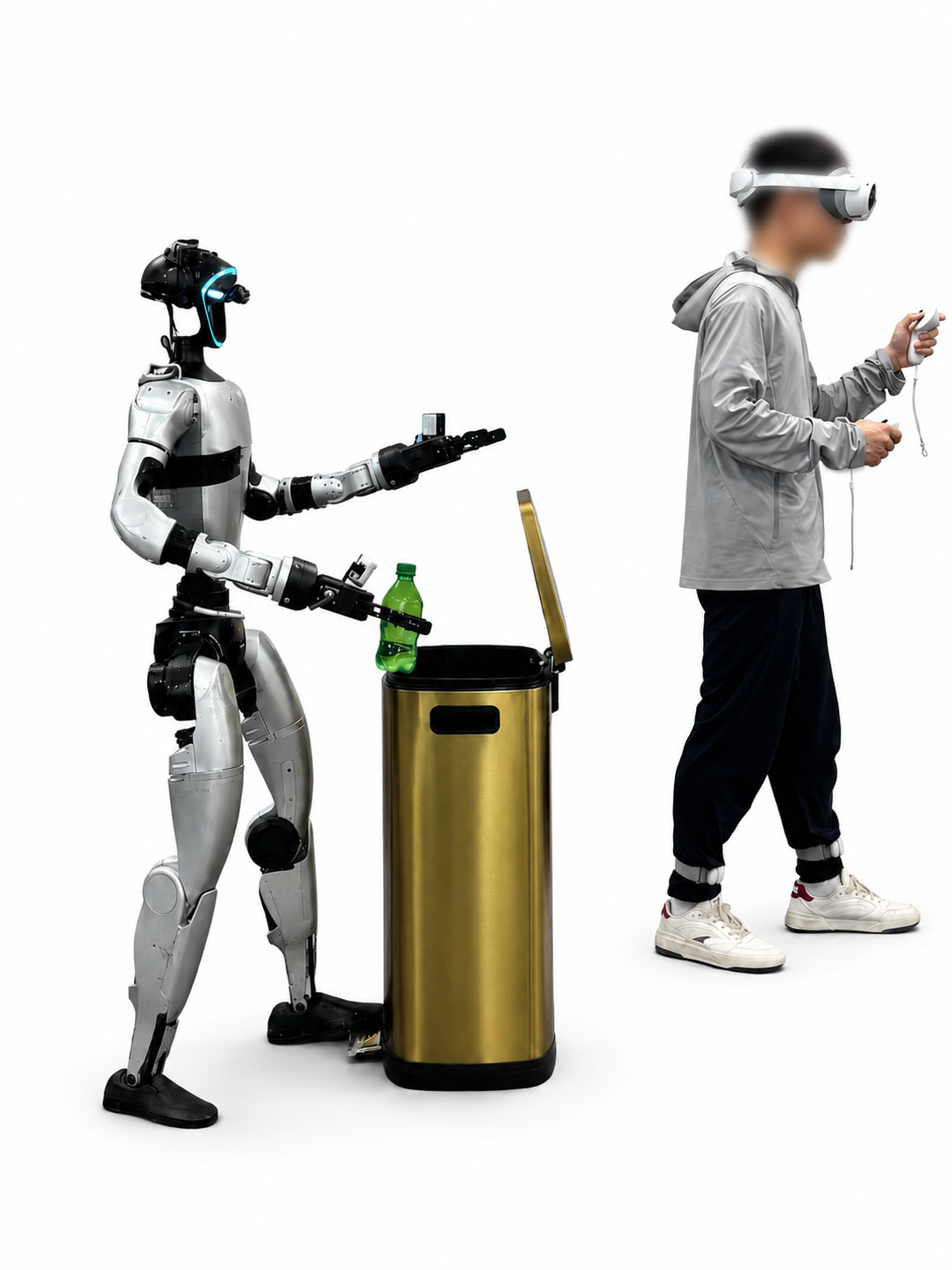}
    \end{minipage}
    \hfill
    \begin{minipage}[c]{0.38\linewidth}
        \centering
        \includegraphics[width=0.88\linewidth]{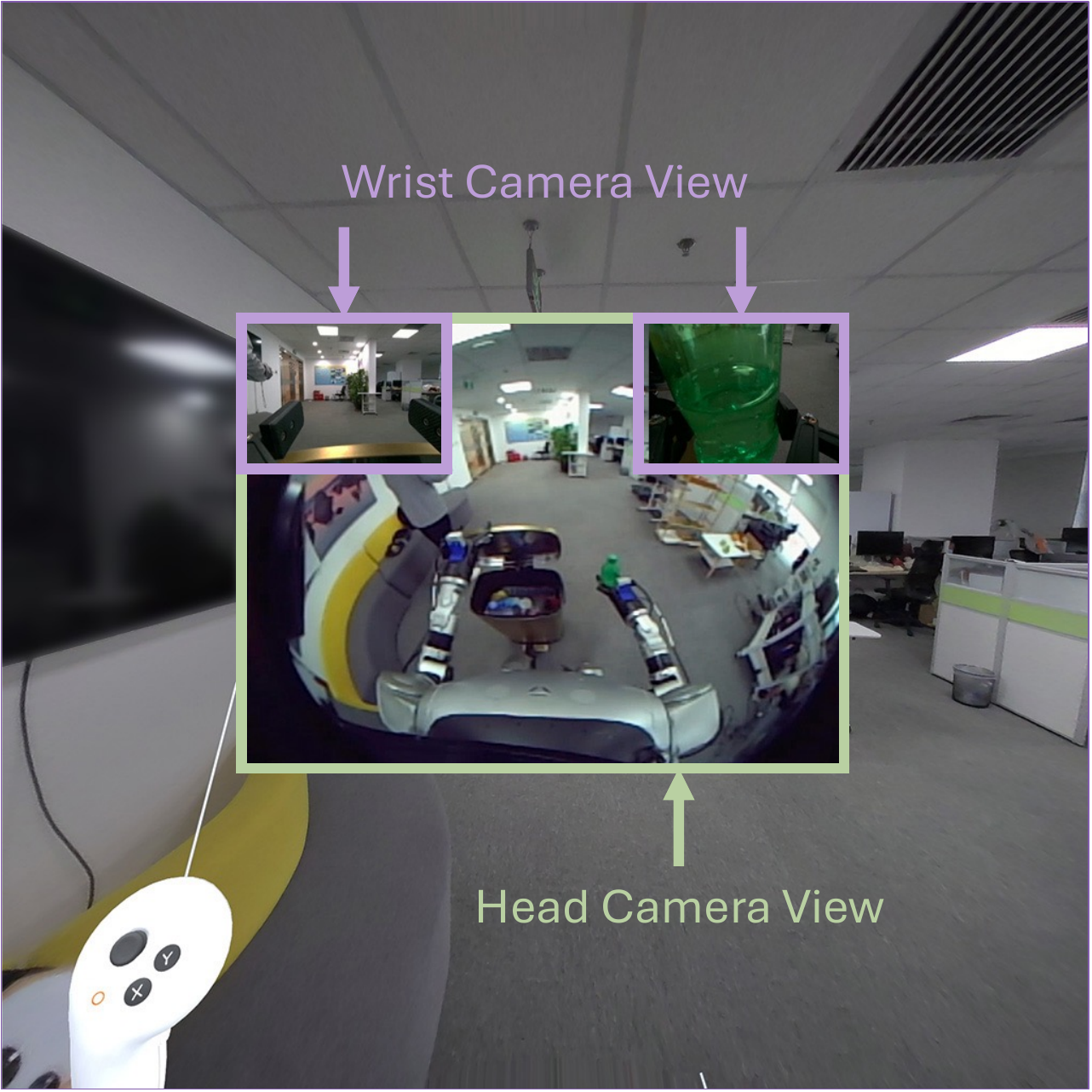}
    \end{minipage}
    \vspace{-1.5em}
    \caption{\textbf{Whole-body teleoperation scene and HMD snapshot, of the same frame.} \textbf{Left}: the PICO4U kit provides the HMD, two handheld controllers, and two leg trackers for live teleoperation. \textbf{Right}: an in-headset egocentric camera view streamed to the operator during teleoperation. The fisheye head view is placed in the center, with the 2 wrist views placed on its top-left and top-right corners. }
    \label{fig:teleop_hardware}
\end{figure}

For all on-robot teleoperation experiments, we use a PICO4U VR kit~\cite{pico2024pico4ultra} consisting of a head-mounted display (HMD), two handheld controllers, and two leg-mounted motion trackers, as shown in Fig.~\ref{fig:teleop_hardware}. The relative poses of these five devices are used to reconstruct the operator's motion, which is later retargeted to the humanoid robot. The HMD streams real-time visual feedback from the robot's head-mounted camera and two wrist-mounted cameras to support egocentric teleoperation, while the triggers on the handheld controllers command the grippers' opening and closing.

\subsection{Teleoperation-Free Data Collection Hardware}

For teleoperation-free data collection, we use the HuMI hardware setup~\cite{nai2026humanoid}. As shown in Fig.~\ref{fig:humi_hardware}, the setup consists of two GoPro-mounted UMI-style handheld grippers~\cite{chi2024universal} and five HTC VIVE Ultimate trackers~\cite{vive_ultimate_tracker}, attached to the two grippers, the pelvis, and the two feet. Note that we adjust the grippers' color and shape to roughly match the robot-mounted grippers; however, differences in camera parameters, camera placement, and gripper opening width leave a domain gap. The system records wrist-view RGB observations, gripper widths, and synchronized 6-DoF tracker poses.

During collection, the demonstrator performs the task using the handheld grippers while a laptop records the synchronized tracker poses. Additionally, an online IK preview visualizes the corresponding humanoid motion in real time, helping the collector adjust demonstrations so that the recorded trajectories remain feasible for the robot.

\section{Implementation Details}
\label{appendix:implementation}

\subsection{Low-Level Controller \& Teleoperation}
\label{appendix:teleop}

This section provides further details on the teleoperation interfaces ablated in \S\ref{sec:roadmap-teleop}. 
All four interfaces use the same visual observations, consisting of stereo RGB images from the head camera and stereo RGB images from the two wrist cameras. 
They also share the same proprioceptive state format: a 32-D body state containing root roll, root pitch, yaw angular velocity, and the 29 measured body joints of the Unitree G1. 
The two grippers are recorded separately as one scalar for each hand. 
The main difference across interfaces is therefore the body-action format: each interface exposes a different command space to the operator, which in turn determines what behaviors can be demonstrated and what action representation the VLA must learn.

\paragraph{Decoupled control teleoperation.}
Decoupled control is widely used in recent humanoid loco-manipulation systems~\cite{cheng2024expressive,liu2024visual,lu2025mobile,ben2025homie,li2025amo,nvidia2025gr00tn16,wei2026psi0}. 
It separates manipulation and locomotion. 
The operator commands the waist and two arms for manipulation, while a lower-body controller receives coarse mobility commands and produces the leg motion. 
This interface is well suited to tasks that can be viewed as mobile manipulation: the robot walks to a workspace, uses its arms, and optionally changes its root height. 
It is less suitable for behaviors that require direct leg or foot participation, because the operator does not command individual lower-body joints.
The recorded body action is 21-D: 3 waist joint targets, 7 left-arm joint targets, 7 right-arm joint targets, 1 base-height command, and a 3-D planar base-velocity command. 
The base-velocity command represents forward velocity, lateral velocity, and yaw angular velocity. 

\paragraph{VR 3-point teleoperation.}
VR 3-point teleoperation follows the sparse head-and-hands interface adopted by prior whole-body tracking systems such as OmniH2O and SONIC~\cite{he2024omnih2o,luo2025sonic}. 
It uses a sparse task-space interface. 
The operator specifies the poses of the two wrists and a head/neck reference point, together with a planar navigation command. 
A motion planner then infers the lower-body motion needed to realize the planar navigation command. 
This format is lightweight and natural for arm-centric tasks, especially reaching and grasping, but it provides only indirect control over torso, knees, and feet. 
The recorded body action is 24-D: a 3-D planar base-velocity command, 9 position coordinates for the three tracked points, and 12 orientation coordinates for the same three points. 
Each tracked point is represented by a 3-D position and a 4-D quaternion, ordered as left wrist, right wrist, and head reference point. 

\paragraph{Joint-based whole-body teleoperation.}
Joint-based whole-body teleoperation follows recent efforts that collect full-body humanoid demonstrations through motion retargeting~\cite{ze2025twist,fu2024humanplus,zhang2025hub}. 
In our implementation, we capture human motion with a PICO VR setup~\cite{pico2024pico4ultra} and retarget it to the robot online using GMR~\cite{araujo2025retargeting}. 
Unlike sparse keypoint interfaces, it gives the operator direct control over the robot's full kinematic chain, including arms, waist, legs, and root motion. 
This makes it suitable for the whole-body behaviors: squatting to expand the reachable workspace, coordinating locomotion with manipulation, and using a foot or leg as an active manipulator. 
The recorded body action is 32-D: root roll, root pitch, yaw angular velocity, and 29 robot joint targets. 
The 29 joint targets consist of 14 arm joints, 12 leg joints, and 3 waist joints, in the Unitree G1 joint order used by the low-level controller. 

\paragraph{SMPL-based whole-body teleoperation.}
SMPL-based whole-body teleoperation uses the SMPL human-body representation~\cite{loper2015smpl}, which is also supported as an input format by recent whole-body tracking controllers~\cite{luo2025sonic}. 
It keeps the action in a human-body representation instead of immediately converting the demonstration into robot joint targets. 
This representation is natural when demonstrations come from human motion capture and is useful for testing whether a human-pose action space can serve as a policy target. 
However, it is substantially higher-dimensional than the robot-joint representation, and many of its coordinates are redundant.
The recorded body action is 81-D: root roll, root pitch, yaw angular velocity, a 6-D wrist/arm refinement term, and 72 SMPL joint coordinates from 24 human-body joints.

\subsection{Whole-Body VLA Policy Design}
\label{appendix:vla}

This section provides further details on the design choices ablated in \S\ref{sec:roadmap-vla} (Fig.~\ref{fig:vla_results}). We organize the discussion into three families: action and proprioception interface, pretraining ablations, and one-step action generation.

\paragraph{Action and proprioception interface.}
We ablate four design choices around the interface between the VLA and the humanoid's action and proprioceptive state. \\
\underline{\textit{Action projection initialization.}}
$\pi_{0.5}$'s action expert contains two linear projection layers: an input projection that maps the noisy action chunk from action-space dimension to the transformer embedding dimension, and an output projection that maps back. The pretrained projections support up to 32 action dimensions, but our whole-body action vector is 34-D (32 dims from \S\ref{sec:roadmap-teleop} plus two parallel-jaw gripper dimensions; a dexterous hand would push this higher), so both projections must be resized.
\textit{Random re-initialization} re-initializes both weight matrices from scratch at the new 34-D shape, discarding the pretrained projection weights entirely.
\textit{Weight surgery} (default) copies the pretrained weights into the first 32 rows/columns of the new matrices and Xavier-initializes only the remaining new entries, preserving the learned action representation for faster, more stable training. \\
\underline{\textit{Action ordering.}}
$\pi_{0.5}$'s pretrained action layout is bimanual, with arms and grippers interleaved per side: [left arm (7), left gripper (1), right arm (7), right gripper (1)]. The humanoid adds 18 dimensions (legs, waist, root pose) that have no counterpart in this layout, so we must decide where to place them.
\textit{Pretrained bimanual ordering with humanoid joints appended} (default) keeps the pretrained slots fixed and concatenates the new dimensions at the end, giving the 34-D vector [left arm (7), left gripper (1), right arm (7), right gripper (1), left leg (6), right leg (6), waist (3), root roll/pitch + yaw rate (3)].
\textit{Humanoid-native ordering} instead places the root and lower body first, [root roll/pitch + yaw rate (3), left leg (6), right leg (6), waist (3), left arm (7), right arm (7), left gripper (1), right gripper (1)]. This is more natural from a humanoid-control perspective but moves the pretrained dimensions to new positions in the action vector. \\
\underline{\textit{Absolute vs.\ relative action targets.}}
For the 29 actuated joints of the Unitree G1 (dual-arm 14 + dual-leg 12 + waist 3), an action chunk admits two parameterizations.
\textit{Absolute targets} (default) expresses each action as absolute joint positions.
\textit{Relative targets} expresses each action in the chunk relative to the first state of that chunk ($a_t - s_0$). \\
\underline{\textit{Proprioceptive input.}}
The humanoid's proprioceptive state is a 34-D vector with the same layout as the action vector.
\textit{With proprioception} (default) feeds this vector into the VLA alongside vision and language, giving the policy direct access to its own body pose, which is otherwise hard to recover since the head- and wrist-mounted cameras do not cleanly observe the lower body.
\textit{Without proprioception} removes this input; the policy must infer body pose from vision alone.

\paragraph{Pretraining ablations.}
We compare three backbone initializations to isolate the contribution of robot pretraining. \\
\underline{\textit{$\pi_{0.5}$ (default).}}
$\pi_{0.5}$~\cite{intelligence2025pi_} is one of the strongest open-source robot VLA models. Architecturally it consists of a pretrained VLM (PaliGemma) plus an action expert that handles robotics-specific inputs and outputs. The full network is pretrained on data from multiple robots, high-level semantic prediction tasks, web data, and other sources to enable broadly generalizable real-world manipulation. Its robot data, however, consists primarily of static or wheeled dual-arm platforms; no humanoid data is included. \\
\underline{\textit{PaliGemma.}}
PaliGemma~\cite{beyer2024paligemma} is the open-source vision-language model on which $\pi_{0.5}$ is built. In this ablation we initialize the VLM portion of the VLA with PaliGemma weights and randomly initialize the action expert, removing all robot-specific pretraining while retaining vision-language representations. \\
\underline{\textit{Random initialization.}}
The entire network (VLM and action expert) is randomly initialized with no pretraining of any kind.

\paragraph{One-step action generation.}
We compare two ways to produce an action chunk in a single forward pass of the action expert. \\
\underline{\textit{One-step flow matching.}}
Training follows the standard flow-matching recipe of $\pi_{0.5}$. At inference, instead of integrating the learned vector field over multiple steps, we integrate from $\tau=0$ to $\tau=1$ in a single step (integration step count set to 1). No retraining is required; this is purely an inference-time change. \\
\underline{\textit{Drifting model.}}
The drifting model~\cite{deng2026generative} is a generative model that learns a pushforward map evolving during training, removing the need for iterative inference and naturally admitting one-step generation. The original paper applies it to robotic control as Drifting Policy, which matches or exceeds the state-of-the-art Diffusion Policy~\cite{chi2025diffusion} (which uses 100-step diffusion-based inference) on single-task visuomotor imitation. We extend it to the multi-task VLA setting by replacing the flow-matching action expert with a drifting-model action expert, keeping the rest of the architecture and the VLM backbone unchanged.

\subsection{Heterogeneous Co-Training}
\label{appendix:heterogeneous_cotraining}

\paragraph{Stationary teleoperation demonstrations.}
Stationary teleoperation uses the same robot and joint-based whole-body interface as our full loco-manipulation demonstrations, but collects only the in-place manipulation portion of a task. During collection, we start after the robot has reached the workspace and stop before it needs to leave it, avoiding the approach and departure motions that require locomotion. The robot therefore stays at the same floor location, while the demonstration still contains arm and gripper manipulation and on-the-spot posture adjustments, such as torso and height changes (e.g., \texttt{Shelf Cube Transfer}) or in-place turns (e.g., \texttt{Pig Placement}). This preserves same-embodiment manipulation data while making collection substantially easier than full loco-manipulation teleoperation, since the operator does not need to teleoperate the robot through the walking portions of each demonstration.

\paragraph{Stationary HuMI demonstrations.}
HuMI collects the same stationary manipulation phases as above, but without the humanoid in the loop, making it faster and cheaper than on-robot teleoperation. After collection, we run offline IK on the recorded tracker trajectories to convert them into whole-body humanoid joint targets. We then package the resulting trajectories in the same state and action format as our joint-based teleoperation data, allowing HuMI episodes to be mixed with teleoperated episodes without any model-side change. For the visual observations, we rectify the GoPro wrist images to roughly reduce the fisheye mismatch, but their wrist views remain noticeably wider than the robot-mounted RealSense views, leaving a substantial visual gap. Finally, to approximate the 0.2\,s future-frame preview latency used in our teleoperation pipeline, we shift the HuMI actions forward by the same amount when forming state-action pairs (with perfect tracking under this latency, a command at the current time corresponds to the robot state 0.2\,s later).

\subsection{Hyperparameters}
\label{appendix:hyperparameters}

Table~\ref{tab:model_hyperparameters} lists the default hyperparameters for training \method. Training on 4 A800 GPUs takes approximately 24 hours. For data augmentation, we apply random crop and random rotation to non-wrist camera observations (head-mounted camera), and random brightness, random contrast, and random saturation to all observation images.

At inference, the VLA generates a 50-step action chunk at 30\,Hz and sends whole-body commands to the low-level SONIC controller. We execute up to 25 actions before running inference again, meaning inference runs every 5/6 seconds.

\begin{table}[h]
\centering
\begin{tabular}{ll}
\toprule
\textbf{Hyperparameter} & \textbf{Value} \\
\midrule
Optimizer & AdamW \\
Optimizer momentum & $\beta_1 = 0.9,\ \beta_2 = 0.95$ \\
Peak learning rate & $1 \times 10^{-4}$ \\
Learning rate schedule & Cosine decay \\
Warmup steps & $1\mathrm{k}$ \\
Training steps & $30\mathrm{k}$ \\
Batch size & $128$ \\
Action horizon & $50$ \\
Observation resolution & $224 \times 224$ \\
\bottomrule
\end{tabular}
\vspace{1em}
\caption{\textbf{Training hyperparameters.} Default configuration for \method VLA training.}
\label{tab:model_hyperparameters}
\end{table}

\newpage
\section{System-Level Comparison on Long-Horizon Tasks}
\subsection{Task Details \& Evaluation Protocol}
\label{appendix:long_horizon}

We provide the detailed settings of the long-horizon task set used for the system-level comparison in Section~\ref{sec:system}.

\begin{taskbox}{Long-horizon Task: \texttt{Fruit Arrangement}}

    \begin{itemize}[leftmargin=4mm]
        \item \textbf{Language Prompt:} Pick the \{fruit 1\} with the right hand, pick the \{fruit 2\} with the left hand, and place them on the shelf.
    
        \item \textbf{Detailed Process:} The robot is required to pick up two designated fruits from a set of distractor fruits using both hands, and then place them into two target containers on the top level of a shelf. Specifically, the robot first picks \{fruit 1\} from a low coffee table at a height of 40 cm using its right hand, and then picks \{fruit 2\} from a medium-height table at a height of 70 cm using its left hand. The target containers are placed on the top level of the shelf at a height of 95 cm. These surfaces span a wide range of heights within the Unitree G1 humanoid's whole-body reachable workspace. The robot must turn and walk to appropriate positions beside each platform before performing the corresponding manipulations, so that the target objects are brought into its reachable range. The robot starts several steps away from the first table and is required to return to the starting location after completing all manipulation steps, forming a closed-loop task sequence for experimental convenience.
        
        \item \textbf{Evaluation Rule:} The task is evaluated with a total of 8 points:
        \begin{itemize}
            \item The robot successfully walks to an appropriate position beside the low coffee table at a height of 40 cm. (1 pt)
            \item The robot kneels down and correctly picks up \{fruit 1\} among the distractor fruits on the low coffee table using its right hand. (1 pt)
            \item The robot stands up and successfully walks to an appropriate position beside the medium-height table at a height of 70 cm. (1 pt)
            \item The robot correctly picks up \{fruit 2\} among the distractor fruits on the medium-height table using its left hand. (1 pt)
            \item The robot turns toward the shelf and adjusts to an appropriate pose for placing the fruits. (1 pt)
            \item The robot raises its right hand above the top level of the shelf and releases \{fruit 1\} into the blue plate. (1 pt)
            \item The robot raises its left hand above the top level of the shelf and releases \{fruit 2\} into the orange basket. (1 pt)
            \item The robot turns around and returns to the starting location. (1 pt)
        \end{itemize}

        \item \textbf{Capability Stressed:} \textit{Long-horizon compositional loco-manipulation}.
    \end{itemize}
\end{taskbox}

The mentioned `fruit 1' and `fruit 2' are chosen from \{banana, peach, mangosteen, lemon, tomato\}, producing $P(5,2)=20$ tasks in total, and during data collection, we collect 6 demos for each pair, with each of their language prompts filled by the specific fruit names, and thus produce 120 demos in total. Except that our \method\,{\scriptsize (HuMI co-training)} adopts only 36 demos (6 pairs) from them and replaces the remaining 84 demos involving held-out fruits (\textit{i.e.}, lemon and tomato) by HuMI demonstrations.

As for evaluation, for each method we test 10 out of 20 pairs that are uniformly chosen as listed in Table~\ref{tab:pairs}. The scene configuration including layout of distractor fruits are recorded to ensure a fair comparison across methods. The performance is evaluated by the average task progress of these 10 rollouts.

\begin{table}[h]
\centering
\resizebox{\columnwidth}{!}{
\begin{tabular}{lcccccccccc}
\toprule
rollout index
& 0 & 1 & 2 & 3 & 4 & 5 & 6 & 7 & 8 & 9 \\
\midrule
fruit 1
& banana & peach & mangosteen & \cellcolor{yellow!25}lemon & \cellcolor{red!15}tomato & banana & peach & mangosteen & \cellcolor{yellow!25}lemon & \cellcolor{red!15}tomato \\

fruit 2
& peach & mangosteen & banana & peach & banana & \cellcolor{yellow!25}lemon & \cellcolor{red!15}tomato & \cellcolor{yellow!25}lemon & \cellcolor{red!15}tomato & mangosteen \\

\bottomrule
\end{tabular}
}
\vspace{0.2em}
\caption{Chosen pairs for long-horizon task evaluation. Each fruit appears as `fruit 1' and `fruit 2' twice. Held-out fruits (\textit{i.e.}, lemon and tomato) of \method\,{\scriptsize (HuMI co-training)} are highlighted.}
\label{tab:pairs}
\vspace{-2em}
\end{table}

\subsection{Baselines \& Data Collection}
\label{appendix:baselines}

\noindent\textbf{GR00T N1.6}~\cite{nvidia2025gr00tn16} is a humanoid VLA built on a Cosmos-2B vision-language backbone~\cite{nvidia2025cosmosreason2} and a diffusion-transformer action head. For the released Unitree G1 loco-manipulation setup, the policy observes an egocentric camera stream and proprioception from the legs, waist, arms, and hands. It outputs a decoupled command at each future step: 14-D relative arm commands, 14-D absolute hand commands, 3-D absolute waist commands, a 1-D base-height command, and a 3-D navigation command. Its non-hand command channels correspond to the 21-D decoupled-control interface in \S\ref{sec:roadmap-teleop} (arms, waist, base height, and navigation), with the difference that GR00T predicts relative rather than absolute arm commands. The low-level controller then executes these commands while handling balance and locomotion. \\
\underline{\textit{Modifications.}}
We preserve GR00T's relative arm representation and the same decoupled body-command interface, but replace the original 14-D hand block with two absolute parallel-gripper width commands for our hardware. We also extend the visual input from the original ego view to three views by adding the left and right wrist cameras. We collect the full loco-manipulation demonstrations for all 20 fruit pairs using this decoupled-control teleoperation setup, yielding 120 demonstrations for fine-tuning. \\
\underline{\textit{Training details.}}
We fine-tune from the official GR00T N1.6 pretrained checkpoint for 50k optimization steps with a global batch size of 32, while keeping the official GR00T fine-tuning recipe, including the optimizer, learning-rate schedule, warmup, and 50-step action horizon.

\noindent$\boldsymbol{\Psi}_0$~\cite{wei2026psi0} is a humanoid VLA that first pretrains a Qwen3-VL~\cite{bai2025qwen3vl} backbone on large-scale egocentric human videos and then post-trains a flow-based MM-DiT action expert on humanoid robot demonstrations. It outputs a 36-D decoupled command at each future step: 14-D hand commands, 14-D arm commands, 3-D torso-orientation commands, a 1-D base-height command, and a 4-D locomotion command consisting of forward velocity, lateral velocity, yaw rate, and target yaw. These commands are executed by AMO~\cite{li2025amo}, an off-the-shelf RL tracking controller. \\
\underline{\textit{Modifications.}}
Using the native $\Psi_0$ pipeline end-to-end would require recollecting the baseline dataset, since its controller interface differs from the GR00T decoupled-control interface used above. To keep the comparison matched while avoiding an additional round of time-consuming data collection, we adapt $\Psi_0$ to the same decoupled-control demonstrations used for GR00T N1.6. We keep the 36-D $\Psi_0$ tensor layout but fill it with our controller fields: the left and right gripper widths occupy one slot in each 7-D hand-command block; the unused hand-action dimensions are masked during training, while the corresponding state dimensions remain zero-padded; the 14-D arm-command block uses the left and right arm commands; the 3-D torso-orientation command and 1-D base-height command use the corresponding controller commands; and the 4-D locomotion command uses our 3-D navigation command, with the target-yaw slot left unused. Thus, the model is trained only on action dimensions present in our controller interface. At deployment, we execute $\Psi_0$ through the same GR00T decoupled-control stack by converting the predicted 36-D action chunks back into the controller command dictionary. We believe the degradation caused by this adaptation has little influence on our conclusions in the main text.\\
\underline{\textit{Training details.}}
We initialize the VLM backbone and action expert from the released $\Psi_0$ pretrained checkpoints, then fine-tune on the same 120 decoupled-control demonstrations for 50k optimization steps with a global batch size of 64. We follow the released $\Psi_0$ fine-tuning recipe, including bf16 training, optimizer settings, 30-step action chunks, and RTC.

\section{Additional Experimental Results}
\label{appendix:additional_exp}

\subsection{Future-frame Preview Latency Analysis}
\label{appendix:latency_gap}
We expand more details of the experiments we conducted to evaluate the effect of different choices of preview latency in this section, as a supplement to the last part in \S~\ref{sec:roadmap-teleop}.

\begin{wrapfigure}{r}{0.6\textwidth}
    \centering
    \vspace{-1.5em}
    \includegraphics[width=\linewidth]{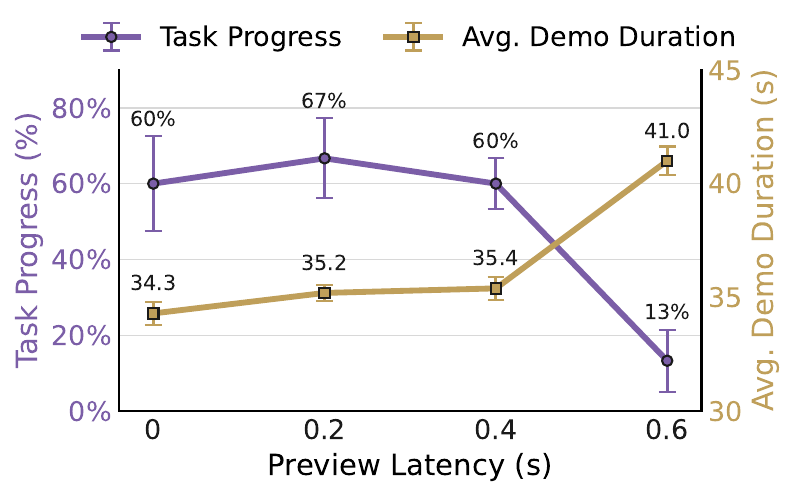}
    \vspace{-1.5em}
    \caption{\footnotesize \textbf{Future-frame preview latency sweep} on \texttt{Cola Placement}.}
    \label{fig:preview}
    \vspace{-1em}
\end{wrapfigure}

We sweep $\Delta t \in \{0, 0.2, 0.4, 0.6\}$\,s collecting the same number of demonstrations (40) per setting on a single task \texttt{Cola Placement}, training one VLA on each, and tracking average demonstration duration as a proxy for teleoperation difficulty. 
Results are shown in Fig.~\ref{fig:preview}.
At $\Delta t = 0$\,s, the robot exhibits the highest responsiveness, thereby delivering the most favorable standing-manipulation experience for the operator. However, this low latency also introduces visible stuttering and ground-stomping during locomotion, during both data collection and testing.
At $\Delta t = 0.6$\,s, accumulated delay overwhelms the operator: demonstration duration jumps from $\sim$35\,s to 41\,s and task progress collapses to 13\%. 
$\Delta t = 0.2$\,s strikes the best balance, reaching 67\% task progress at a demonstration duration (35.2\,s) essentially unchanged from the zero-preview case. 
This finding holds across multiple teleoperators in our pipeline, and $\Delta t = 0.2$\,s yields high-quality demonstrations on all subsequent tasks. A better low-level motion tracker that relies less on future frames may further reduce the optimal value of this $\Delta t$ in the future.

Note that on the same \texttt{Cola Placement} task and with the same preview latency $\Delta t=0.2$\,s, the peak performance 67\% showed in Fig.~\ref{fig:preview} is much lower than the 87\% progress of the same task presented before in Table~\ref{tab:teleop} and Fig.~\ref{fig:smpl}, because previous models are trained on multiple tasks, specifically, the first 4 tasks of our HLM-12 benchmark. Here in Fig.\ref{fig:preview} in \S~\ref{sec:roadmap-teleop}, all models are trained on a single task to ablate the preview latency only. This performance gap suggests that multi-task mixed training outperforms single-task training, potentially by enabling the enhancement of partially overlapping skills, such as navigating to the target furniture or locating and grasping objects.

\subsection{Scaling of Teleoperation Demonstrations}
\label{appendix:scaling}
\begin{wrapfigure}{r}{0.6\textwidth}
    \centering
    \vspace{-1em}
    \includegraphics[width=\linewidth]{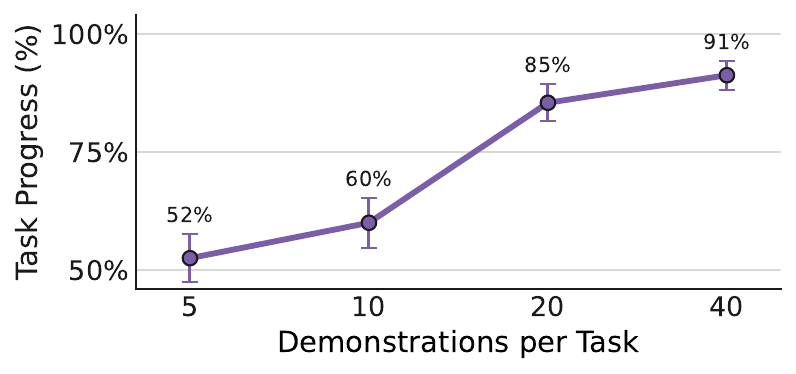}
    \caption{\footnotesize \textbf{Whole-body teleop data scaling.}}
    \vspace{-0.5em}
    \label{fig:data_ratio}
    \vspace{-1em}
\end{wrapfigure}

With the VLA architecture settled at the end of \S~\ref{sec:roadmap-vla}, we care about the amount of teleop data it demands to reach a satisfactory performance. We sweep 4 values (5, 10, 20, 40) on the 4-task subset (first 4 tasks of HLM-12) results are shown in Fig.~\ref{fig:data_ratio}. The largest jump comes between 10 and 20 demos per task; returns flatten thereafter, reaching roughly 90\% at 40 demos. 40 demonstrations is a modest budget, costing a skilled operator about 1.5 hour per medium-difficulty task; every ablation up to this point already uses it as the default whole-body teleop budget, and we will keep this amount for later experiments unless explicitly noted.

\subsection{Per-Task Results on the HLM-12 Benchmark}
\label{appendix:per_task}

\begin{figure}[h]
    \centering
    \includegraphics[width=1.0\linewidth]{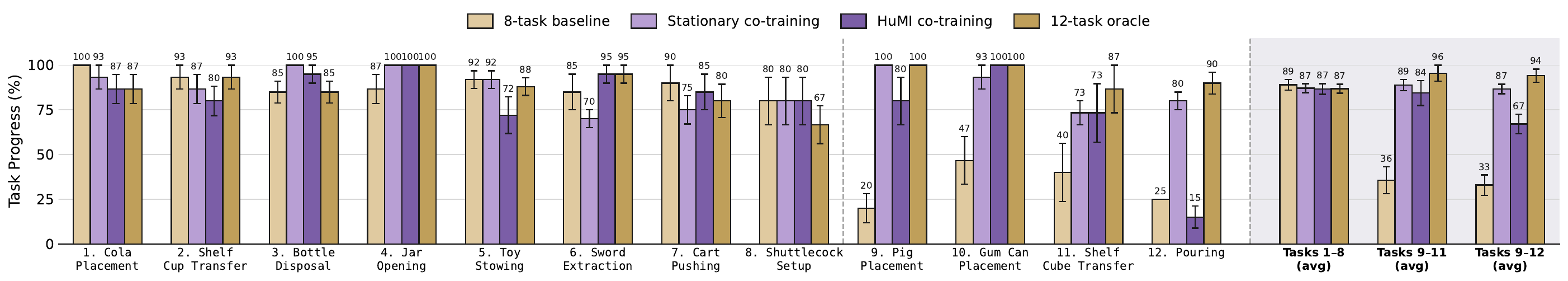}
    \caption{\footnotesize \textbf{Per-task breakdown of heterogeneous co-training results.} Task progress for all 12 tasks and three aggregates (rightmost). Four conditions: 8-task baseline, stationary co-training, HuMI co-training, and 12-task oracle. Co-training does not regress training tasks. On held-out tasks, both methods reach near-oracle on motion-reuse tasks (9--11); stationary succeeds on the new-motion task (12), HuMI does not.}
    \label{fig:co_training_per_task}
\end{figure}

Figure~\ref{fig:co_training_per_task} shows per-task results on the HLM-12 benchmark for the four conditions evaluated in \S\ref{sec:roadmap-cotrain}. All 12 tasks appear individually, plus three aggregate bars (rightmost): Tasks 1--8 (training), Tasks 9--11 (motion-reuse held-out), and Tasks 9--12 (all held-out). Four conditions are compared: 8-task baseline, stationary co-training, HuMI co-training, and 12-task oracle.

On training tasks (1--8), all methods maintain similar performance; co-training does not regress the base policy. On held-out tasks (9--12), the gap is stark. For motion-reuse tasks (9--11), both co-training methods reach near-oracle performance, supplying new semantic understanding (new objects and prompts). On Task 12 (\texttt{Pouring}), which requires a new motion (vessel tilt), stationary co-training matches the oracle while HuMI co-training fails, consistent with the visual and action gaps discussed in \S\ref{sec:roadmap-cotrain}.

\subsection{Scaling HuMI Demonstrations}
\label{appendix:humi_scaling}

\begin{wrapfigure}{r}{0.6\textwidth}
    \centering
    \includegraphics[width=\linewidth]{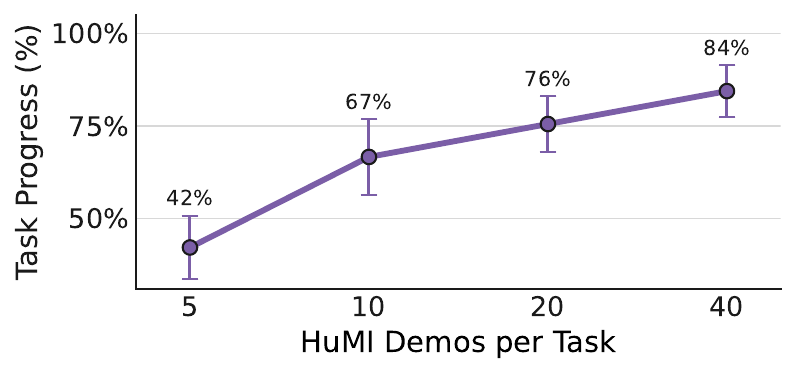}
    \caption{\footnotesize \textbf{Scaling HuMI demonstrations per task.} Average task progress on the three motion-reuse held-out tasks (Tasks 9--11) as we vary the number of HuMI demonstrations per task, with the 8-task whole-body teleop set fixed. Performance climbs from 42\% at 5 demos to 84\% at 40 demos.}
    \label{fig:humi_data_ratio}
\end{wrapfigure}

Figure~\ref{fig:humi_data_ratio} examines how HuMI demonstration count affects performance on the three motion-reuse held-out tasks (Tasks 9--11), with the 8-task whole-body teleop set fixed.

Performance scales from 42\% at 5 demos to 67\% at 10, 76\% at 20, and 84\% at 40. The largest gain comes between 5 and 10 demos (+25 points); returns diminish thereafter (+9 from 10 to 20, +8 from 20 to 40). Even 5 demonstrations per task provide substantial signal, confirming HuMI's effectiveness at supplying semantic understanding for motion-reuse tasks. These experiments focus on motion-reuse tasks where HuMI already succeeds; scaling HuMI data to enable motion transfer (e.g., Task 12, \texttt{Pouring}) is an interesting direction for future work that we plan to explore.

\end{document}